\pdfoutput=1
\documentclass[11pt,a4paper]{article}
\usepackage[T1]{fontenc}
\usepackage[utf8]{inputenc}
\usepackage{textcomp}
\usepackage{amsmath,amssymb}
\usepackage{newtxtext,newtxmath}

\makeatletter
\newcommand{\microsign}{{\usefont{TS1}{qtm}{\f@series}{\f@shape}\symbol{181}}}
\makeatother
\usepackage{newunicodechar}
\newunicodechar{°}{\textdegree}
\newunicodechar{±}{\textpm}
\newunicodechar{µ}{\microsign}
\newunicodechar{ũ}{\~{u}}
\newunicodechar{Δ}{\ensuremath{\Delta}}
\newunicodechar{α}{\ensuremath{\upalpha}}
\newunicodechar{β}{\ensuremath{\upbeta}}
\newunicodechar{ε}{\ensuremath{\upvarepsilon}}
\newunicodechar{η}{\ensuremath{\upeta}}
\newunicodechar{λ}{\ensuremath{\uplambda}}
\newunicodechar{μ}{\ensuremath{\upmu}}
\newunicodechar{π}{\ensuremath{\uppi}}
\newunicodechar{ρ}{\ensuremath{\uprho}}
\newunicodechar{σ}{\ensuremath{\upsigma}}
\newunicodechar{τ}{\ensuremath{\uptau}}
\usepackage[margin=2.4cm]{geometry}
\usepackage{graphicx,booktabs,tabularx,array,calc,longtable}
\IfFileExists{footnotehyper.sty}{\usepackage{footnotehyper}}{\usepackage{footnote}}
\makesavenoteenv{longtable}
\usepackage{etoolbox}
\usepackage{placeins}
\usepackage{caption}
\DeclareCaptionLabelSeparator{pipe}{\ $|$\ }
\usepackage{microtype}
\usepackage{setspace}
\usepackage{xcolor}
\definecolor{brand}{HTML}{0E9C7E}\definecolor{brandink}{HTML}{0B5E50}\definecolor{brandbg}{HTML}{F1FBF8}
\usepackage{xurl}
\usepackage[colorlinks=true,linkcolor=brandink,citecolor=brandink,urlcolor=brandink]{hyperref}
\usepackage{titlesec}
\titleformat{\section}{\sffamily\large\bfseries\color{brandink}}{\thesection}{0.6em}{}
\titleformat{\subsection}{\sffamily\normalsize\bfseries\color{brandink}}{\thesubsection}{0.6em}{}
\titleformat{\subsubsection}{\normalfont\normalsize\bfseries}{\thesubsubsection}{0.6em}{}
\titlespacing*{\subsubsection}{0pt}{1.0ex plus .3ex}{0.2ex}
\titlespacing*{\section}{0pt}{1.6ex plus .4ex}{0.8ex}
\titlespacing*{\subsection}{0pt}{1.2ex plus .3ex}{0.5ex}
\usepackage{fancyhdr}
\newsavebox{\absbox}
\renewenvironment{abstract}{\par\noindent\setlength{\fboxsep}{9pt}\setlength{\fboxrule}{0.8pt}\begin{lrbox}{\absbox}\begin{minipage}{\dimexpr\linewidth-2\fboxsep-2\fboxrule\relax}\vspace{0.4ex}\centerline{\sffamily\large\bfseries\color{brandink}\abstractname}\vspace{0.6ex}\small\setstretch{1.08}\noindent\ignorespaces}{\par\vspace{0.4ex}\end{minipage}\end{lrbox}\fcolorbox{brand}{brandbg}{\usebox{\absbox}}\par\vspace{1.2em}}
\providecommand{\tightlist}{\setlength{\itemsep}{0pt}\setlength{\parskip}{0pt}}
\makeatletter
\patchcmd\longtable{\par}{\if@noskipsec\mbox{}\fi\par}{}{}
\newcommand\needspacefor[1]{\par\penalty-100\begingroup\setlength{\dimen@}{#1}\dimen@ii\pagegoal\advance\dimen@ii-\pagetotal\ifdim\dimen@>\dimen@ii\ifdim\dimen@ii>\z@\vfil\fi\break\fi\endgroup}
\newsavebox\pandoc@box
\newcommand*\pandocbounded[1]{\sbox\pandoc@box{#1}\Gscale@div\@tempa{\textheight}{\dimexpr\ht\pandoc@box+\dp\pandoc@box\relax}\Gscale@div\@tempb{\linewidth}{\wd\pandoc@box}\ifdim\@tempb\p@<\@tempa\p@\let\@tempa\@tempb\fi\ifdim\@tempa\p@<\p@\scalebox{\@tempa}{\usebox\pandoc@box}\else\usebox{\pandoc@box}\fi}
\def\fps@figure{!htbp}
\makeatother
\usepackage[normalem]{ulem}

\renewcommand{\abstractname}{Abstract}

\hypersetup{pdftitle={EnzymeRSI: A dual-loop recursive self-improvement system for enzyme design},pdfauthor={Yuejie Hou, Lixiang Yang, Yisi Liu, Weiting Tang, Yueyuxiao Yang, Meng Yang}}

\usepackage{titling}
\pretitle{%
  \noindent\includegraphics[height=15mm,alt={Genoria AI logo}]{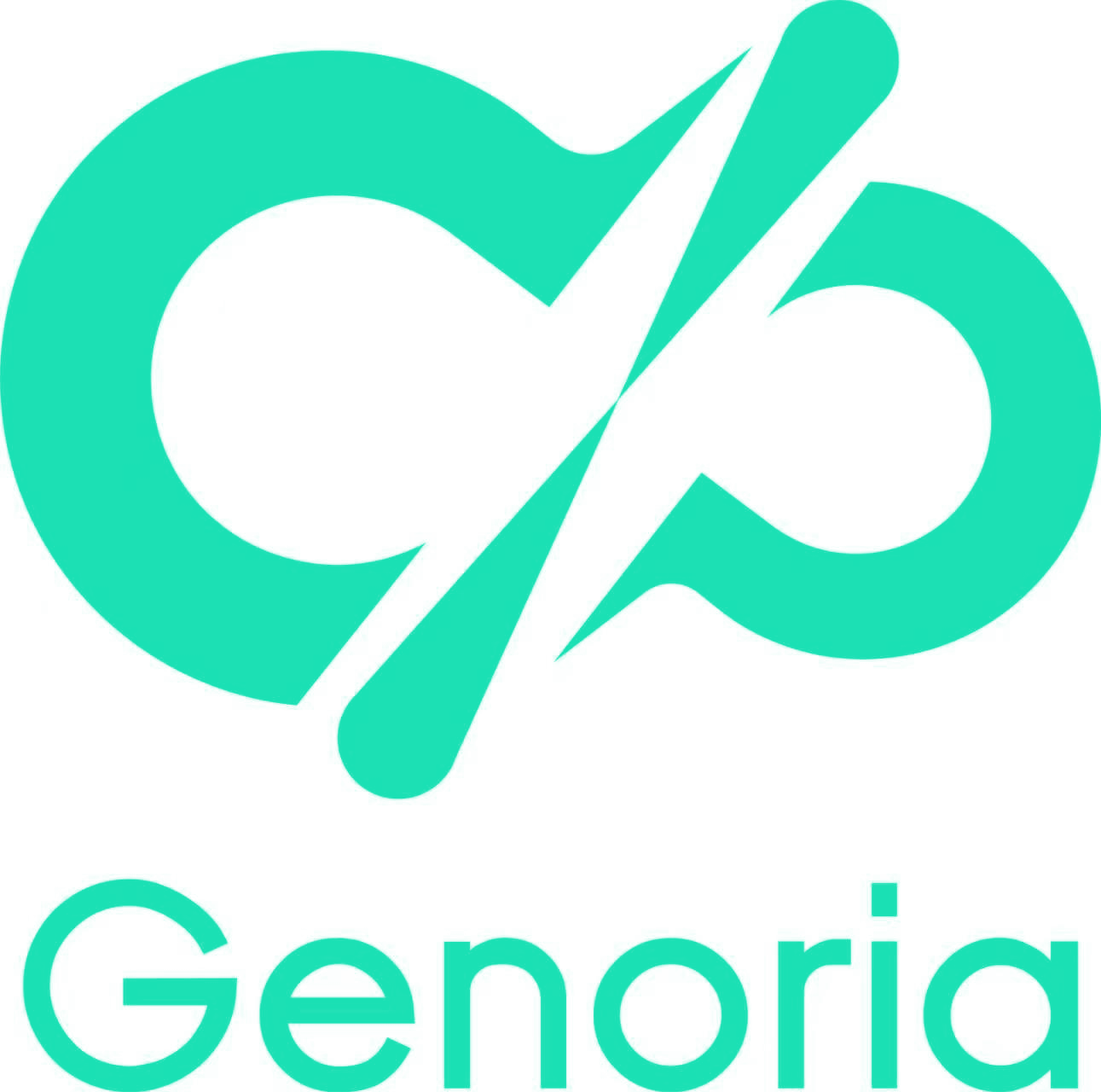}\hfill\begin{tabular}[b]{@{}r@{}}{\sffamily\Huge\bfseries\color{brand}EnzymeRSI}\\[0.3ex]{\sffamily\small\color{brandink}Dual loop $\cdot$ Recursive self-improvement}\end{tabular}\par
  \vspace{0.8ex}\noindent{\color{black}\rule{\linewidth}{1.6pt}}\par\vspace{1.4ex}
  \begingroup\centering\sffamily\LARGE\bfseries\setstretch{1.05}}
\posttitle{\par\endgroup\vspace{1.0ex}%
  {\centering\noindent{\color{black!60}\rule{0.9\linewidth}{0.4pt}}\par}\vspace{1.2ex}}
\preauthor{\begingroup\centering\normalfont}
\postauthor{\par\endgroup}
\predate{\begingroup\vspace{0.3em}\centering\small\color{black!65}}
\postdate{\par\endgroup}

\date{4 November 2026}
\title{EnzymeRSI:\\A dual-loop recursive self-improvement system for enzyme~design}
\author{%
{\large\sffamily Genoria AI Team}\\[0.5ex]
{\normalsize\sffamily Yuejie Hou, Lixiang Yang, Yisi Liu, Weiting Tang, Yueyuxiao Yang, Meng Yang\textsuperscript{*}}\\[0.4ex]
{\normalsize\sffamily Genoria AI, Shenzhen 518083, China}\\[0.4ex]
{\small\sffamily\color{black!70}*Correspondence to: yangmeng@genoria.ai}}
\begin{document}

\thispagestyle{plain}

\maketitle\vspace{-0.2em}

\begin{abstract}Improvement in enzyme engineering is usually pursued by optimising which candidates to assay within a fixed model and workflow; when the bottleneck arises from a mismatched method or a failing workflow, that route is difficult to escape. We present EnzymeRSI, a dual-loop system for recursive self-improvement (RSI). The outer loop evolves harness assets without changing model weights, the inner loop develops prediction and selection algorithms under controlled conditions, and a cloud laboratory system is specified for closing the physical loop with quality-controlled feedback. By updating its prediction strategy through iterative comparison, the outer loop raised the mean development-set Spearman correlation on ProteinGym by 0.044 across five assays, reaching 0.335 on CD19 and 0.683 on OPSD (rhodopsin); of the two first places, only CD19 was produced by the outer loop itself. In the self-improvement of the ThermoMPNN algorithm the inner loop reached a PCC of 0.849, and on TadA-Bench it reached a Spearman correlation of 0.307 in 1.13 h. We examined the system's experimental decision-making in three enzyme systems. For the luciferase RLuc, 10 of the 32 nominated substitutions that had not been pre-specified reached at least twice the wild-type activity in a single round of measurement. For the cytidine deaminase A3A, a variant selected by a rule that caps loss of unmodified C raised 5mC conversion in the weaker CT context from 11.4\% to 46.5\% and reversed the substrate preference in both contexts. For the DNA polymerase KOD-E, 189 single-substitution variants defined a landscape on which different objective functions selected divergent next steps from the same parent, and one assayed combination reached 31.1 times the parent on a modified substrate. By treating the organisation of the research workflow and the development of algorithms as separately scored objects of evolution, and linking a frozen decision record to a physical measurement, this work offers an evolutionary paradigm for moving beyond the blind spots of a fixed method and for autonomous enzyme engineering.\end{abstract}

\phantomsection
\section{1 Introduction}\label{ux524dux8a00}

Directed evolution turned mutation, screening and amplification into a repeatable engineering workflow [1], and machine-learning-guided directed evolution went further by using measured data to decide which candidates to assay next [2]. How many samples a round of wet-lab work can process, and how long that round takes, determine how often a method can be tested, and therefore how much judgement itself is worth: choosing the right thing to measure matters more than measuring more.

Computational capability has changed by orders of magnitude in recent years. AlphaFold brought structure prediction to experimentally useful accuracy [3], protein language models can infer structure and fitness directly from sequence [4], ProteinMPNN made backbone-to-sequence design routine [5], and deep learning can support engineering even in the few-shot regime [6]. The range of feasible designs has widened accordingly: a natural-language description can drive \emph{de novo} design [7], generative models can be steered towards properties that natural proteins do not possess [8], iterative fine-tuning around catalytic motifs can explore complex active sites [9], design pipelines for reactions and non-natural substrates that nature has never evolved are now open source [10], and structure generation and reasoning have begun to alternate within a single chain [11]. The matching evaluation resources have matured in parallel: ProteinGym assembles large-scale mutational effect measurements [12], stability measurements at the million scale provide a common training base [13], and transfer learning on those data has further improved the prediction of stability changes [14].

Active-learning-assisted directed evolution decides the next batch to assay from uncertainty and expected gain [15], protein-language-model-driven rapid \emph{in vitro} evolution obtains marked improvement within a few rounds [16], and methods that model epistasis explicitly show that the substitution to add next depends on the current sequence context [17]. These three lines of work have defined the field in recent years, and they share one assumption: that the method itself is fixed. They optimise which candidates to assay, not the method used to judge which candidates to assay.

The same foundation model performs very differently under different forms of organisation. With the foundation model frozen, improvements in evidence fusion and invocation configuration alone can improve fitness prediction [18]. A protein design harness built around standardised tool interfaces is one example: across nine biomolecular design tasks it reduced token consumption by 86\% and GPU cost by 64\% while raising the computational design pass rate, and in one zero-shot end-to-end design it yielded a low-nanomolar cortisol binder [19]. Structure and affinity prediction have likewise been integrated into a unified drug design engine [20], agent-driven discovery engines orchestrate folding, affinity, design and decision together [21], and open-source all-atom engines place folding, reasoning and scale-up in one pipeline [22]. What these systems change is not model weights but prompts, workflows, tool access and subtask division, that is, the harness. Together they indicate that the harness delivers substantial gains and yet has long gone unmanaged as an object of evaluation in its own right.

The research process itself is also being automated. Automated laboratories can already complete multiple rounds of protein evolution in a robotic closed loop without human intervention [23], and mobile robots can operate instruments and run experiments unattended for long periods [24]. Decisions made computationally can therefore be traced through to real measurements. Agents that execute research actions have also appeared: large-model systems can plan and execute chemical experiments autonomously [25] or complete multi-step chemical reasoning through tool use [26]. On the biological side, autonomous agents have completed experimentally validated \emph{de novo} drug design, in which designs against human IL-6 passed quality control, entered wet-lab testing and yielded nanomolar binders [27]. Other work has tested directly whether coding agents can build biological AI tools themselves [28], and has used agent-guided design for epitope removal [29].

Method modules can also be improved automatically. Program search has produced mathematical discoveries [30] and automated research workflows can produce papers end to end [31]; general algorithm discovery agents [32] and sample-efficient program evolution frameworks [33] have turned code improvement into a repeatable loop, and research-and-development agents for the full data science workflow have taken shape [34]. In molecular discovery, systems have used Monte Carlo tree search over the space of scientific workflows, combined with persistent cross-task memory, and have achieved a comparable improvement in wet-lab hit rate relative to expert-designed workflows [35]. The harness itself has also been treated as an object of recursive self-improvement: one line of work evolves the harness without labels and uses Pareto selection to trade task performance against execution cost, raising the pass rate on coding benchmarks by about 2.0 percentage points and lowering per-task cost by about 26\% relative to earlier methods [36]. The harness has further begun to be dissected and consolidated systematically, both through source-level study of several coding-agent harnesses [37] and through harnesses that learn from execution experience [38].

What this work lacks is a comparable basis for evaluation. Existing benchmarks mostly measure task completion, for example how agents perform on machine-learning engineering tasks [39], rather than the improvement itself: whether it has been independently evaluated, whether it was actually adopted, and whether it can be reused in later tasks. Comparisons between frameworks are also often built on single runs, different foundation models and different budgets, which cannot separate a system that has genuinely improved from the same system run twice with a better second result. Existing harness self-improvement work takes the pass rate and cost of coding benchmarks as its endpoint [36] and has not yet faced wet-lab work, where real measurements carry batch noise and quality-control status, and where the cost of adopting a change is a round of experiments rather than a re-run of a program.

We therefore built EnzymeRSI, a dual-loop recursive self-improvement (RSI) system that connects a dry-side decision to a wet-side measurement, scoring the organisation of the research workflow and the development of algorithms as separate objects, with quality-controlled feedback. The outer loop evolves harness assets without changing the weights of the foundation model, updating tool routing and prediction strategy through iterative comparison against a fixed control. The inner loop develops prediction and selection algorithms under controlled conditions and on demand, and connects to real physical measurement and quality-control feedback through the cloud laboratory system. In computational evaluation, outer-loop evolution raised the mean development-set Spearman correlation on ProteinGym by 0.044 across five assays, reaching 0.335 on CD19 and 0.683 on OPSD (rhodopsin); in the self-improvement of the ThermoMPNN algorithm the inner loop reached a PCC of 0.849, and in multi-round extrapolation on TadA-Bench it reached a Spearman correlation of 0.307 in 1.13 h. In three wet-lab enzyme engineering tests, the system addressed cold-start nomination, a dual-substrate constraint trade-off, and multi-step path selection in evolution. Cold-start nomination for RLuc identified 10 substitutions with at least twice the wild-type activity among the 32 candidates that had not been pre-specified. For A3A, Y130Q emerged as the top-ranked variant under a 50\% cap on unmodified-C loss, reversed the substrate preference in both sequence contexts, and raised 5mC conversion in the weaker context from 11.4\% to 46.5\%. For KOD-E, 189 single-substitution variants defined a landscape on which two objective functions led to different next steps from the same parent, and one assayed combination reached 31.1 times the parent; the contact surfaces in the crystal structure accounted for the physical mechanism of the reversal in rank between positions 676 and 680. This work therefore shows that treating research organisation rules and scientific algorithms together as verifiable objects of evolution can move past the blind spots of a fixed method, and provides an evolutionary paradigm for autonomous enzyme engineering.

\phantomsection
\section{2 Results}\label{ux7ed3ux679c}

\phantomsection
\subsection{2.1 A dual-loop architecture scores research organisation and method development separately}\label{ux53ccux73afux67b6ux6784ux628aux7814ux7a76ux7ec4ux7ec7ux4e0eux65b9ux6cd5ux7814ux53d1ux5206ux5f00ux8ba1ux5206}

EnzymeRSI represents an enzyme design task as \(\text{q}=\left( \text{x}_{\text{0}},\text{g},\text{c},\text{D}_{\text{vis}},\text{B} \right)\): the parent sequence, the objective together with the constraints that may not be violated, the protein and experimental context, the data the system may legitimately access, and a multidimensional budget (Fig. 1). The improvement space is partitioned explicitly into two loops. The outer loop (Harness RSI) performs non-parametric version evolution over prompts, pipelines, skills and subagents, and does not modify model weights. The inner loop (Algorithm RSI) is triggered on demand, and only after engineering and workflow causes have been excluded, for method bottlenecks that existing tools cannot resolve; it develops predictors, training recipes, search operators and acquisition functions under controlled conditions. Candidates from both loops are scored in dry-lab evaluation. ProtoPilot [41] then orchestrates the wet-lab protocol and delegates high-frequency decisions, including candidate pre-screening, batch admission and quality control (QC), to the Operon-S1 decision model [43]. Once the execution layer has completed the physical measurement, the results return as Evidence carrying QC status and as a research Trajectory. The evaluation layer scores protein performance, algorithmic capability and system execution on three independent tiers.

\begin{figure}[!htbp]
\centering
\includegraphics[width=5.69in,height=0.62\textheight,keepaspectratio]{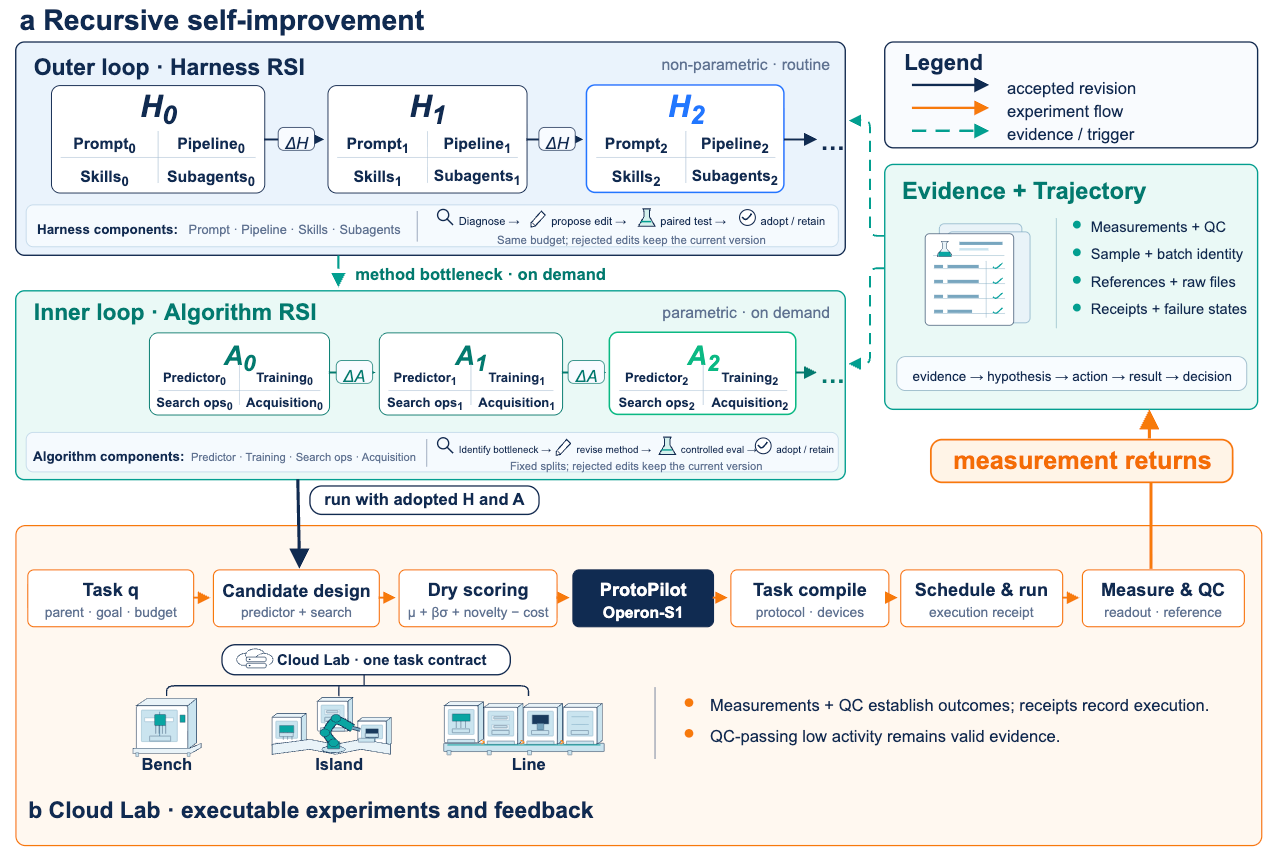}
\caption*{\textbf{Fig. 1 $|$ The dual-loop structure of EnzymeRSI.} \textbf{a}, Dual-loop recursive self-improvement. The outer loop evolves versions of prompts, pipelines, skills and subagents, which are adopted or rolled back after diagnosis, a minimal consistent change bundle and a paired comparison at matched budget; model weights are not modified. The inner loop is triggered only when a method bottleneck appears, and develops predictors, training strategies, search operators and acquisition functions under controlled conditions. Missing packages and permission problems are handled as engineering faults and trigger neither loop. Adoption rules are fixed before the comparison as global adoption, scoped retention or rollback, and an adopted version must be loaded by the next run. The protein, algorithm and system tiers are scored independently and are not collapsed into a single weighted number (Section 4.7). Evidence and Trajectory record measurements, QC status, sample identity, batch, references, checksums of raw files and execution receipts, and return along the sequence evidence, hypothesis, action, result, decision. They may trigger either loop or neither. The legend distinguishes three arrow types: a black arrow is an accepted revision, an orange arrow is experiment flow and a green dashed arrow is evidence or a trigger, which is used as required. \textbf{b}, Cloud Lab. Candidates that have been scored in dry-lab evaluation and passed ProtoPilot--Operon-S1 fast decisions form a wet-lab commission, which runs on the adopted version, is orchestrated by ProtoPilot and compiled into device tasks, and is then scheduled, measured and quality-controlled by the execution layer. The same task contract can address a single instrument, a workcell, a production line or a cloud laboratory. A receipt evidences a software event only; an actual transfer requires measurement and QC support. A sample that passes QC but shows low activity is valid evidence, whereas a failed run enters execution diagnosis. Measurements return to Evidence and Trajectory. The cloud-laboratory path was exercised for protocol execution and measurement in the three enzyme cases; these experiments did not constitute a multi-round autonomous RSI campaign.}
\end{figure}

\phantomsection
\subsection{2.2 Inner loop: algorithm RSI}\label{ux5185ux73afux7b97ux6cd5-rsi}

The inner loop was exercised on two dry-lab tasks drawn from enzyme engineering. The ThermoMPNN stability task predicts the effect of a single substitution on folding stability and is used to judge whether a substitution will destabilise the protein. The TadA amino-acid view requires the activity ranking of variants in a later round to be predicted from completed rounds of directed evolution, which reproduces the situation faced when a campaign is advanced.

On ThermoMPNN, EnzymeRSI reached a development-set Pearson correlation coefficient (PCC) of 0.849, higher than all ten reference frameworks, and the gain coincided with the deliberate introduction of ESM2 representations (Fig. 2a, b). Only two of the ten reference frameworks moved in that direction: MLEvolve used the 35M and 150M variants and stopped at 0.813, whereas OpenRSI, the highest of the ten, left the representation unchanged and reached 0.834 by ensembling the official model with nine fully fine-tuned members (Table 2). For some frameworks the valid candidate output was an exact affine transform of the official prediction. Because the Pearson correlation is invariant to a positive affine transform, the development score then remained identical to the starting point, which appears in Fig. 2a as a curve lying on the starting line. Twelve runs were completed on this task; the final state of each reference framework is given in Table 2.

On the TadA amino-acid view, the development-set Spearman correlation reached 0.307, and the same 20 formal evaluations were completed in about half the active time (1.13 h versus 2.24 h; Fig. 2c). On this task the training data and the evaluation objective are not aligned. By round 28 the variable positions had collapsed from 75 to 28 and were concentrated in the C-terminal region 119--166, and the median activity was 0.025 in the training set but 1.03 in the development set. The supervision available from the early rounds therefore indicates which substitutions inactivate the enzyme, whereas the candidates to be ranked have all survived. EnzymeRSI replaced continuous regression with ordinal modelling over activity bins, so that the model fits activity rank rather than activity value; QwenCode, which migrated across more than ten model architectures, stopped at 0.301.

\begin{figure}[!htbp]
\centering
\includegraphics[width=6.00in,height=0.62\textheight,keepaspectratio]{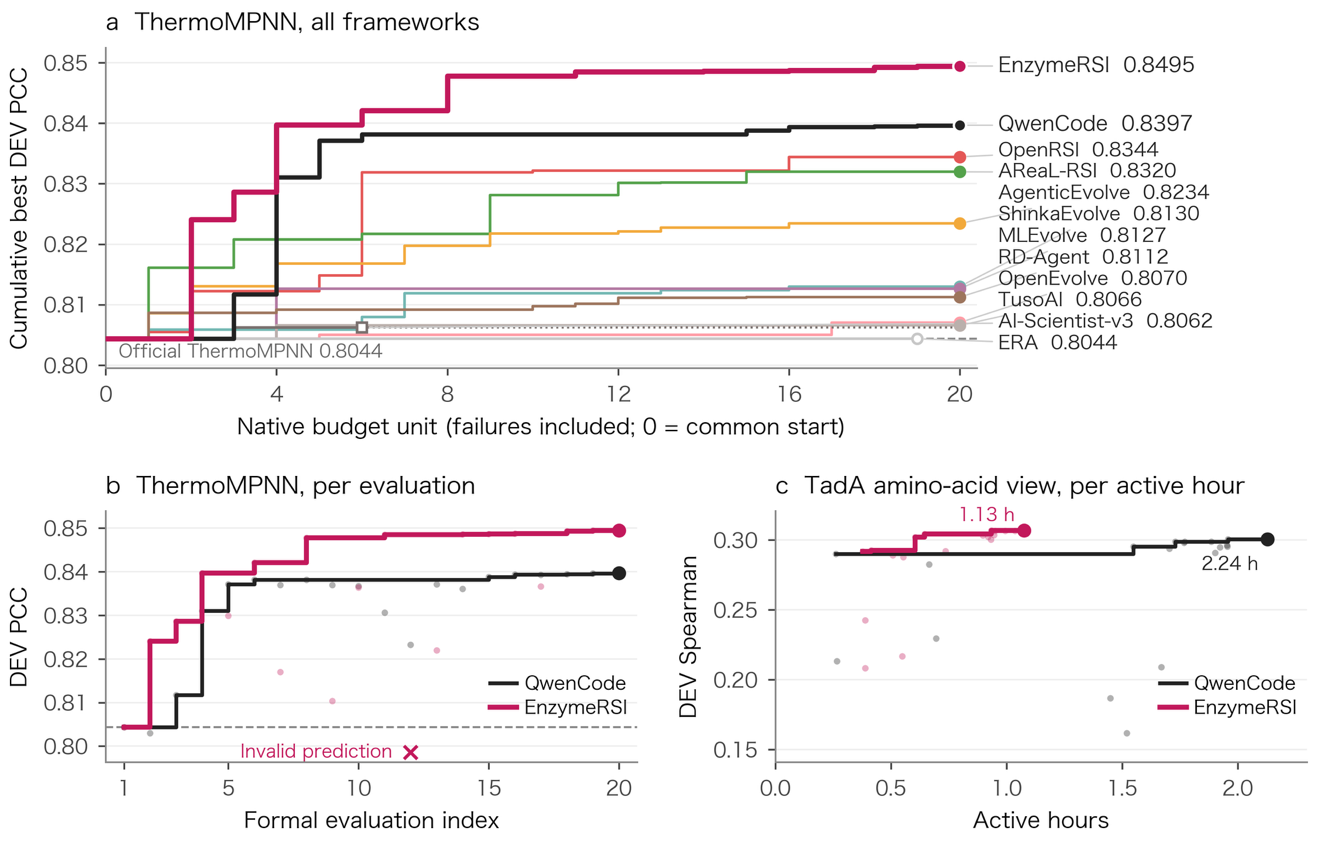}
\caption*{\textbf{Fig. 2 $|$ EnzymeRSI and the reference frameworks on two development tasks.} \textbf{a}, Cumulative-best trajectory on the ThermoMPNN development set. The horizontal axis counts the framework's own outer-loop rounds, each framework on its own scale, and includes failed evaluations; 0 is the common starting point. Each framework ran its native number of rounds and evaluations; horizontal position is therefore not a comparison of compute. The vertical axis is the cumulative best development-set PCC. The dashed line marks the official ThermoMPNN predictor. Filled circles denote a run that completed its native number of rounds, open circles one that did not, and squares an early stop. EnzymeRSI, QwenCode and OpenRSI, the highest of the reference frameworks, are labelled at the end of their curves; the remaining reference frameworks are labelled directly. \textbf{b}, Evaluation-by-evaluation view on ThermoMPNN, with the formal evaluation index on the horizontal axis. Step lines are the cumulative best, pale points are individual evaluations, crosses below the axis are submissions with invalid predictions that were not scored, and the dashed line is the official starting point. \textbf{c}, Evaluation-by-evaluation view on the TadA amino-acid view, with cumulative active hours on the horizontal axis, which is the quantity reported as ``Wall clock'' in Table 3, and the development-set Spearman correlation on the vertical axis; step lines and pale points are as in \textbf{b}. The two curves have the same number of formal evaluations, 20, and each curve ends at the time of the last formal evaluation. The QwenCode curves in \textbf{a} and \textbf{b} are from the unmodified run. Full values and run details for every system are given in Tables 2 and 3.}
\end{figure}

\phantomsection
\subsection{2.3 Outer loop: the self-harness agent}\label{ux5916ux73af-self-harness-agent}

The outer loop was exercised on five human protein assays, one assay per ProteinGym functional category; the category label printed in each panel of Fig. 3 is given first in parentheses. Folding stability (Stability, OTU7A) predicts whether a single substitution causes the domain to unfold. Antibody binding (Binding, CD19) requires 3,761 variants to be ranked by whether they are still recognised by a monoclonal antibody. Expression of rhodopsin (Expression, OPSD) judges whether the protein reaches its correct location. Cell viability (Activity, NPC1) and channel function (Organismal fitness, SCN5A/Nav1.5) ask whether the protein still works in a cellular and in a whole-physiological context. The outer loop changes only how the harness is organised, which includes the inputs that are read, the order in which tools are called and the points at which checks are placed; it does not change model weights.

When the five endpoints were placed among the 97 models in the official ProteinGym record, binding and expression ranked first and viability third. Binding was the only first place obtained by the evolution itself: the starting point of 0.314 was level with the leading model, MIF, at 0.316, and two adoptions over 39 rounds raised it to 0.335, with the gain concentrated in a single change to the variant input found at round 12 (Fig. 3). Our reproduction of MIF differed from the official record by no more than 0.0004, so this margin does not arise from measurement bias. Expression also ranked first, but its starting point of 0.680 already exceeded the leading value of 0.656 and 44 rounds contributed only 0.003; its first place therefore derives from the tool combination used at the starting point rather than from the evolution. Across the five assays, the development-set Spearman correlation rose by an average of 0.044 relative to the respective starting points (Fig. 3).

\begin{figure}[!htbp]
\centering
\includegraphics[width=6.00in,height=0.62\textheight,keepaspectratio]{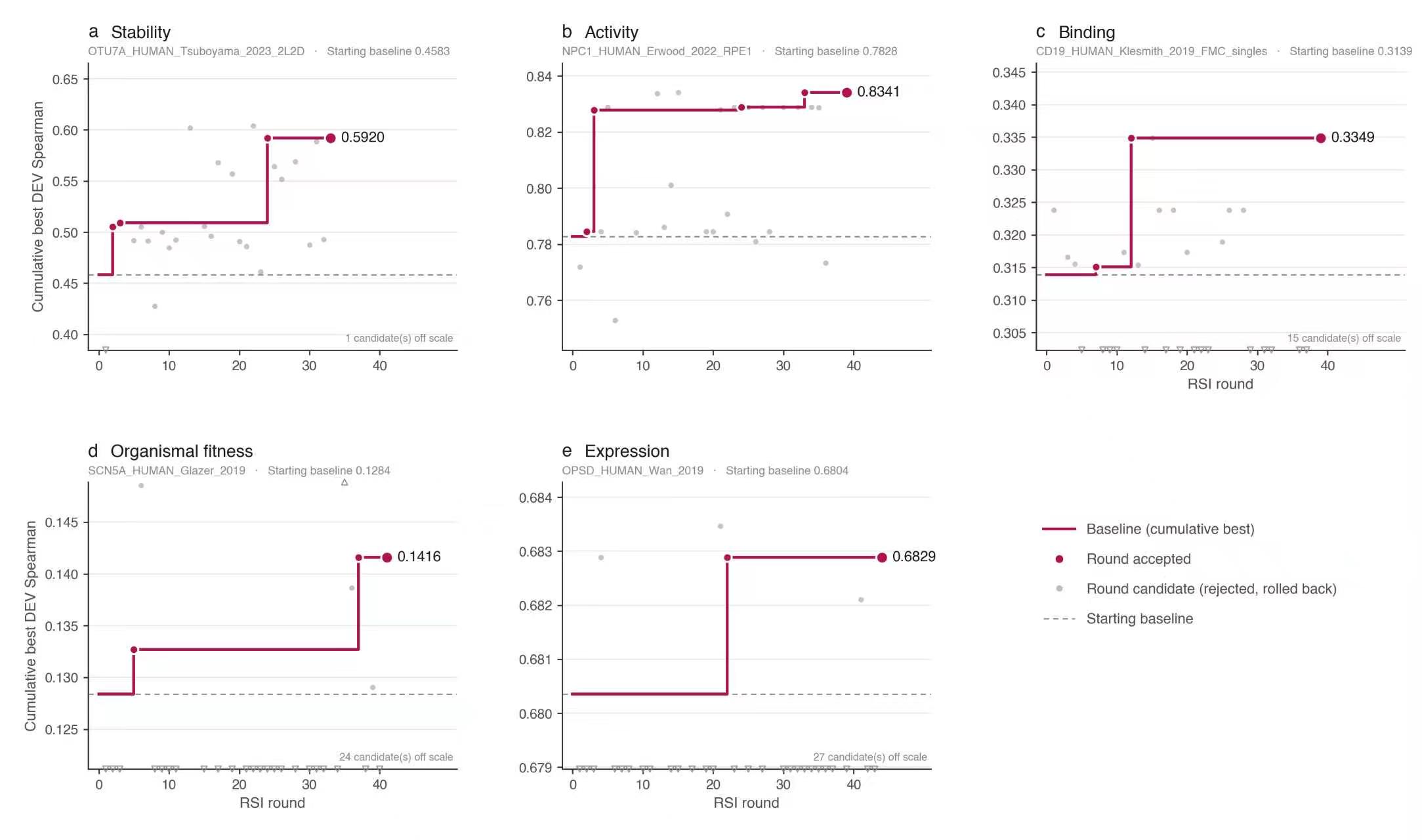}
\caption*{\textbf{Fig. 3 $|$ Outer-loop self-improvement trajectories on five classes of protein measurement.} The 196 rounds are shown in one panel per property, each making paired comparisons on a single assay: \textbf{a}, folding stability; \textbf{b}, intracellular function; \textbf{c}, binding to the FMC63 monoclonal antibody, where the panel ranks 3,761 variants; \textbf{d}, sodium-channel variant function; \textbf{e}, expression and localisation. The horizontal axis is the RSI round and the vertical axis is the cumulative best Spearman correlation on the development set of that assay. The dark magenta step line is the baseline, which rises only in a round that passes both the paired comparison and the gate. Filled circles mark adopted rounds; the filled circle at the end of each line is the endpoint, followed by the final value and the gain relative to the starting point. Grey points are the candidates of that round, of which those that did not pass have been rolled back. The grey dashed line is the starting baseline. Each panel subtitle gives the ProteinGym record identifier for that assay together with the value of the starting baseline, from which the endpoint and the gain can be recomputed panel by panel. Candidates falling outside the range of the vertical axis are drawn as open triangles on the corresponding side, and their number is stated inside the panel. Across the whole run there were 142 paired comparisons and 12 adoptions (8.5\%); the primary metric was spearman\_macro with a minimum acceptable increment of min\_delta = 0.001.}
\end{figure}

\phantomsection
\subsection{2.4 Three enzymes test single substitutions, constraint-dependent variant switching, and path selection under epistasis}\label{ux4e09ux9176ux6e7fux5b9eux9a8cux5206ux522bux9a8cux8bc1ux5355ux70b9ux7a81ux53d8ux53ccux7ea6ux675fux53d8ux4f53ux5207ux6362ux4e0eux4e0aux4f4dux6548ux5e94ux4e0bux7684ux8defux5f84ux9009ux62e9}

The computational evaluation above examined how EnzymeRSI improves methods and the organisation of research. We next examined its support for experimental decisions in three enzyme systems, which differ in what the system is asked to do and in the evidence available to it. On RLuc, the system combined literature, sequence and structural information to nominate single-substitution candidates before the measurements of this round, and the measurement then examined both how the candidates performed and how they were ranked within the pool. On A3A, the system used the measured data from both sequence contexts to turn the requirement to retain unmodified C and the objective of 5mC conversion into screening and ranking rules. On KOD-E, the system used the measured variants to compare the subsequent substitutions and the path selection that follow from different existing mutants under different engineering objectives. The three cases concern candidate nomination, constrained selection and path decision respectively, and differ in the nature of the evidence available: RLuc is a prospective nomination frozen before measurement, whereas A3A and KOD-E recompile rules on data already measured. The steps for which the system was responsible and the corresponding experimental basis are given in Table 5.

\textbf{RLuc. Three candidates nominated on inference alone and frozen before measurement each exceeded the wild type, but the same experiment exposed weak ranking within the pool.} Before the luminescence measurements of this round were obtained, EnzymeRSI combined single-point measurements from the literature with sequence and structural information and nominated 40 single substitutions of RLuc covering 31 positions; the design record and the evidence grading were frozen before the experiment. Eight of these were pre-specified RLuc8 components. Among the remaining 32 candidates that had not been pre-specified, 16 gave a higher integrated readout than the same-plate wild type and 10 reached at least twice the wild-type value; for all 40 candidates the corresponding counts were 22 and 13 (Fig. 4).

Three nominations are the most informative. M185I, S287I and S287M rested on homology, structure or computational inference alone, with no single-point support from the literature. Their integrated readouts across three independent preparations were 2.30, 1.98 and 2.43 times the wild type; all three exceeded the wild type, and M185I and S287M reached at least twofold. Noise in the measurement layer affects the relative fold directly: the coefficient of variation of the wild-type reference-well integrals was 25.6\%, 65.6\% and 44.7\% on the three plates, and removing a single reference well from the second plate multiplied the relative fold of every candidate on the plate by 0.84 to 1.61. We therefore recomputed the results of these three candidates while removing each same-plate wild-type reference well in turn. Under the three removals, the mean integrated readout of S287M was 3.91, 2.04 and 2.04 times the wild type and so remained above twofold throughout. M185I and S287I rose to 3.70 and 3.18 times when the high-value reference well F5 on plate 2 was removed, and fell to 1.93 and 1.66 times when either low-value reference well was removed. The finding that all three exceed the wild type therefore does not depend on any single reference well, whereas the twofold threshold is sensitive to the composition of the reference wells and only S287M is robust to it. This also shows, from the measurement side, that returned Evidence must carry QC status and the composition of the reference set rather than a single scalar reading.

The same experiment exposed a boundary of the method: the design rank of a candidate and its measured rank were only weakly correlated (Spearman \(\text{ρ}\) = 0.175). The hits of this round therefore came from the composition of the candidate pool, whereas ranking within the pool carried almost no information. This is direct evidence that the algorithm layer requires better calibration and ranking, and it is one of the conditions under which the inner loop is triggered.

\begin{figure}[!htbp]
\centering
\includegraphics[width=5.83in,height=0.62\textheight,keepaspectratio]{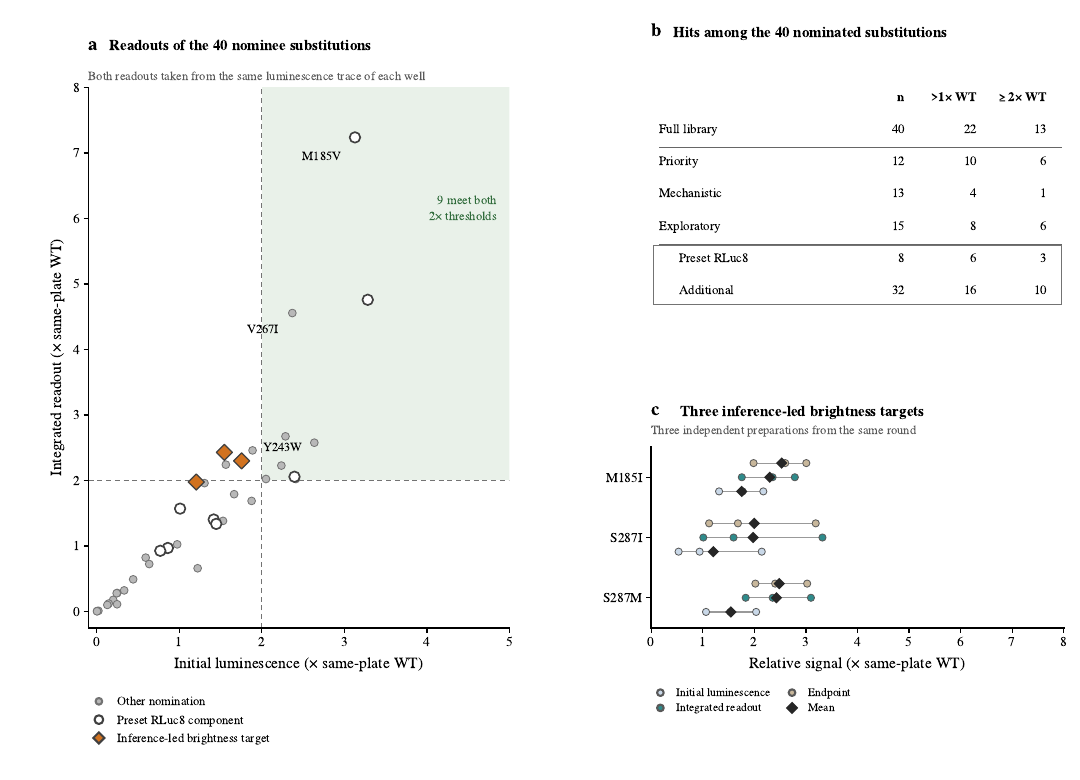}
\caption*{\textbf{Fig. 4 $|$ Pre-experiment single-point nominations for RLuc and their luminescence readouts.} \textbf{a}, The 40 nominated single substitutions. Both readouts are taken from the same luminescence trace of each well and normalised to the same-plate wild type: the horizontal axis is the initial luminescence and the vertical axis is the integrated readout over the first 290.2 s. Dashed lines mark twice the wild-type value; the shaded region contains the nine substitutions that exceed both. Diamonds mark the three brightness candidates that rest on inference alone, and open circles the eight pre-specified RLuc8 components. \textbf{b}, Hit counts for the 40 nominations, grouped by the design category assigned before the experiment; a hit is a relative signal above the wild type ($>$1$\times$ WT) or at least twice it ($\geq$2$\times$ WT). \textbf{c}, Initial, integrated and endpoint readouts of the three inference-led candidates, from three independent preparations in the same round; diamonds mark the mean of the three.}
\end{figure}

\textbf{A3A. The top-ranked variant differed at three limits on unmodified-C conversion, and substrate preference was reversed in both sequence contexts.} The engineering objective was to deaminate 5-methylcytosine (5mC) preferentially in both the CG and the CT sequence contexts while limiting the bypass loss of unmodified cytosine (C). From the requirement set by the researchers, namely to cap the C conversion rate and to maximise the 5mC conversion rate in the weaker context, EnzymeRSI compiled an executable two-step selection rule: filter an eligible set using the higher of the two C conversion rates as a hard constraint, then rank by the lower of the two 5mC conversion rates. The rule was generated and applied on the same set of 75 measured variants; the analysis therefore concerns how the optimum moves with the constraint, not how well an unmeasured variant would be ranked. Applied to these 75 variants with complete paired measurements, the selected variant differed at each of three constraint settings (the full threshold frontier in Fig. 5a contains six top choices and five switches). At a C conversion limit of 50\%, 18 variants were eligible and the system selected Y130Q, whose 5mC conversion in the weaker context was 46.51\%, against 4.55\% for the runner-up F66K. When the limit was relaxed to 60\%, the eligible set expanded to 24 and the top choice became K47T, at 64.61\%. When at least 90\% of unmodified C had to be retained in both contexts, that is, a C conversion limit of 10\%, only 6 variants were eligible and the top choice became Y132C, at 3.11\%. On the same measured data the optimum therefore jumps discretely as the boundary condition changes, which indicates that a dual-objective constraint must enter the adoption operator explicitly and cannot be folded into a single weighted score in advance.

The analysis of the measured data by EnzymeRSI further resolved the magnitude and the structural origin of the reversal in substrate preference. The wild type converted unmodified C preferentially in both contexts, at 92.27\% against 69.41\% for C and 5mC respectively in the CG context and 34.81\% against 11.43\% in the CT context. Y130Q reversed this preference in both contexts, to 46.44\% against 90.19\% in CG and 6.07\% against 46.51\% in CT (Fig. 5c). Of the 75 paired variants, 9 converted 5mC more efficiently than C in both contexts, and the positive preference difference of Y130Q, that is 5mC minus C, still reached 40.44 percentage points on the weaker side (Fig. 5b). By contrast, K47T raised the lowest 5mC conversion rate to 64.61\% but also raised its loss of unmodified C in the CT context to 39.01\%, so that the two variants suit strict and permissive requirements for C retention respectively. Using the crystal structure of A3A in complex with single-stranded DNA containing unmodified cytosine (5KEG) [40], EnzymeRSI localised this phenotype to the core of the active-site pocket in a geometric calculation that used no activity labels: the shortest heavy-atom distance between the Tyr130 aromatic ring and the C5 atom of the target cytosine is only 3.66 \AA, which offers a direct account of the sensitivity of substitutions at this position to the C5 methylation state.

\begin{figure}[!htbp]
\centering
\includegraphics[width=5.83in,height=0.62\textheight,keepaspectratio]{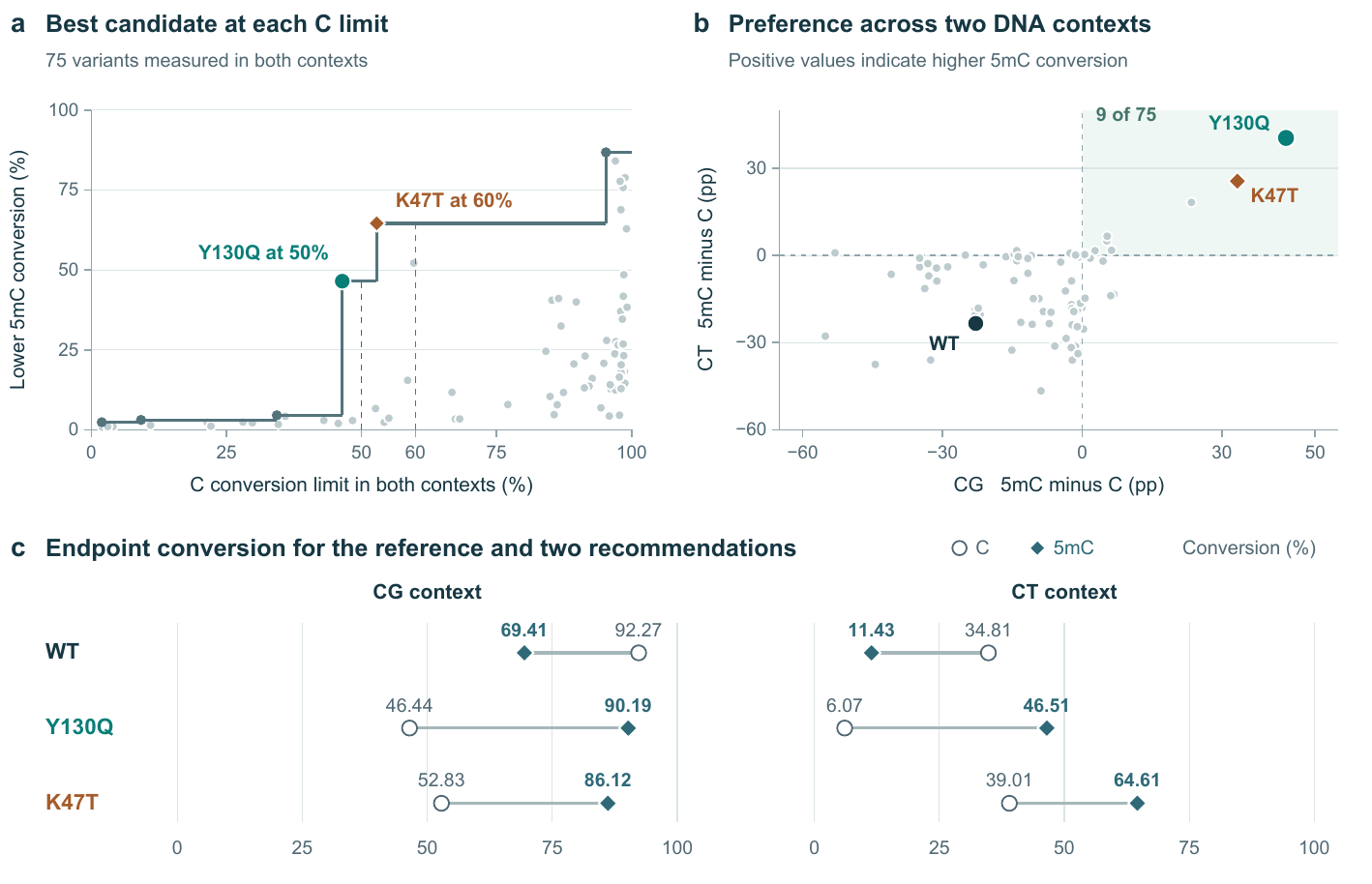}
\caption*{\textbf{Fig. 5 $|$ Dual substrate objectives and the variant selected by the A3A rule.} \textbf{a}, Highest attainable value of the lower 5mC conversion rate across the two contexts, given a limit on the C conversion rate in both contexts; the variants selected at the 50\% and 60\% limits are labelled. \textbf{b}, Preference difference, 5mC minus C, within the CG and CT contexts for 75 variants; the upper right quadrant holds the 9 variants that favour 5mC in both contexts. \textbf{c}, The unmodified-C and 5mC conversion endpoints for the wild type, Y130Q and K47T, each measured in the CG and the CT context.}
\end{figure}

\textbf{KOD-E. Path selection in evolution: different objective functions lead to divergent routes, and positions in close contact produce pronounced reversals of rank.} With the engineered archaeal polymerase KOD-E as the parent and its relative catalytic efficiency set to 1.0, the 189 variants were assayed for the incorporation efficiency of Cy3-modified dATP; these efficiencies form the landscape on which subsequent additions and alternative paths are compared. The system then paired those constructs that differ by a single additional substitution into 125 parent--descendant pairs, covering 41 mutant parents (Fig. 6a). Among the 19 parents with at least two assayed descendants, 10 carried branches that both raised and lowered efficiency. For example, among the 45 variants of E485I, which itself reaches 2.9-fold, the further addition of S451L raised this to 31.1-fold whereas the addition of R484G lowered it to 1.2-fold (Fig. 6b). On this landscape, selecting a route by different objective functions led EnzymeRSI to markedly different next-step choices (Fig. 6c). Starting from the single mutant Q461A at 1.3-fold, an objective of the highest efficiency among direct descendants selected the addition of E485I, reaching 3.8-fold, whereas an objective of the longest strictly monotonic path selected the addition of L457M, reaching 1.5-fold and continuing through K465Q and D718Q to 1.6-fold and 1.8-fold. Critically, when each of E485I at 3.8-fold and L457M at 1.5-fold was completed with the substitution of the other, the two routes converged on the same triple mutant, QH307 at 2.6-fold. The final step towards an identical endpoint sequence therefore appears as a gain in activity along one path, from 1.5-fold to 2.6-fold, and as a loss along the other, from 3.8-fold to 2.6-fold. This shows that the substitution to recommend depends strongly on how the engineering objective is defined, and that the outer loop must compile the researcher's objective function explicitly into selection and stopping rules.

The paired analysis, together with a structural analysis performed without access to construct activity values, further resolved the spatial mechanism of epistasis. Among the 82 additional substitutions covered by the 125 single-step comparisons, 18 occurred in at least two sequence backgrounds, and 8 of these reversed the direction of their effect between backgrounds; 7 still did so when both sides were required to differ by more than 0.1 parent units (Fig. 7a). Among the 21 sets of four matched constructs in which the remainder of the sequence was fixed, 7 showed a reversal of rank, and all 7 involved positions 676 and 680 (Fig. 7b, c). In the parental Met680 background, His at position 676 was markedly better than Ser, at 3.0-fold against 1.1-fold, a difference of \(-1.9\) parent units for Ser minus His. After position 680 was mutated to Ile, Ser overtook His, at 9.4-fold against 3.9-fold, so that the difference reversed to \(+5.5\) and the swing across backgrounds reached 7.4 parent units. Adding E485I lowered both to 2.4-fold and 1.5-fold and narrowed the gap to 0.9. Without being given any construct activity value, EnzymeRSI performed geometric calculations for the position pairs specified by the researchers on the KOD ternary complex (5OMF) and the binary structure (4K8Z) (Fig. 8). Positions 676 and 680, which show the pronounced reversal of rank, lie only 3.9 \AA{} apart at the closest heavy atoms, or 4.0 \AA{} in 4K8Z, and approach the template and primer strands directly at 3.4 \AA{} and 5.0 \AA{} respectively. By contrast, positions 451 and 485, whose combination gives the 31.1-fold activity, are 14.4--16.7 \AA{} apart, and position 485 lies 8.6 \AA{} from the template strand and 8.9 \AA{} from the incoming dATP, in both cases outside the range of direct contact. This spatial contrast indicates that strong coupling between positions in close direct contact favours sign epistasis, whereas distant positions can deliver large gains through conformational cooperation without direct contact.

\begin{figure}[!htbp]
\centering
\includegraphics[width=0.86\linewidth,height=0.62\textheight,keepaspectratio]{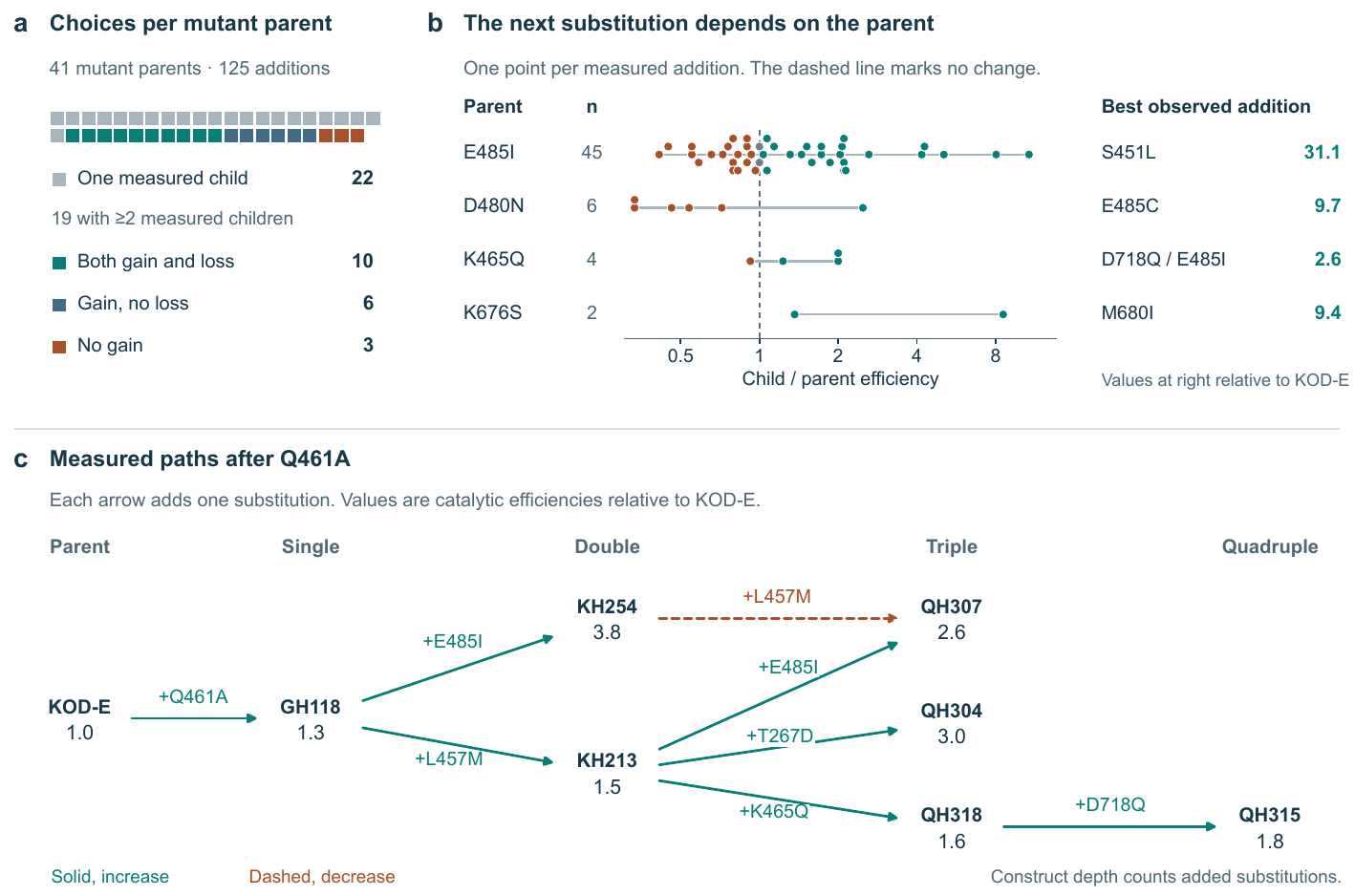}
\caption*{\textbf{Fig. 6 $|$ Single-step selection and evolutionary paths for KOD-E.} \textbf{a}, Distribution of parents with one or more assayed descendants, classified by whether gains and losses both occur among the descendants. \textbf{b}, All comparable descendants of four parents, E485I, D480N, K465Q and K676S; the horizontal axis is the efficiency ratio of the descendant to its direct parent, and the right-hand column gives the best assayed additional substitution for each parent. \textbf{c}, Assayed paths from Q461A, where level denotes the number of substitutions, solid lines are gain edges, dashed lines are loss edges, and values are catalytic efficiencies relative to KOD-E.}
\end{figure}

\begin{figure}[!htbp]
\centering
\includegraphics[width=0.86\linewidth,height=0.62\textheight,keepaspectratio]{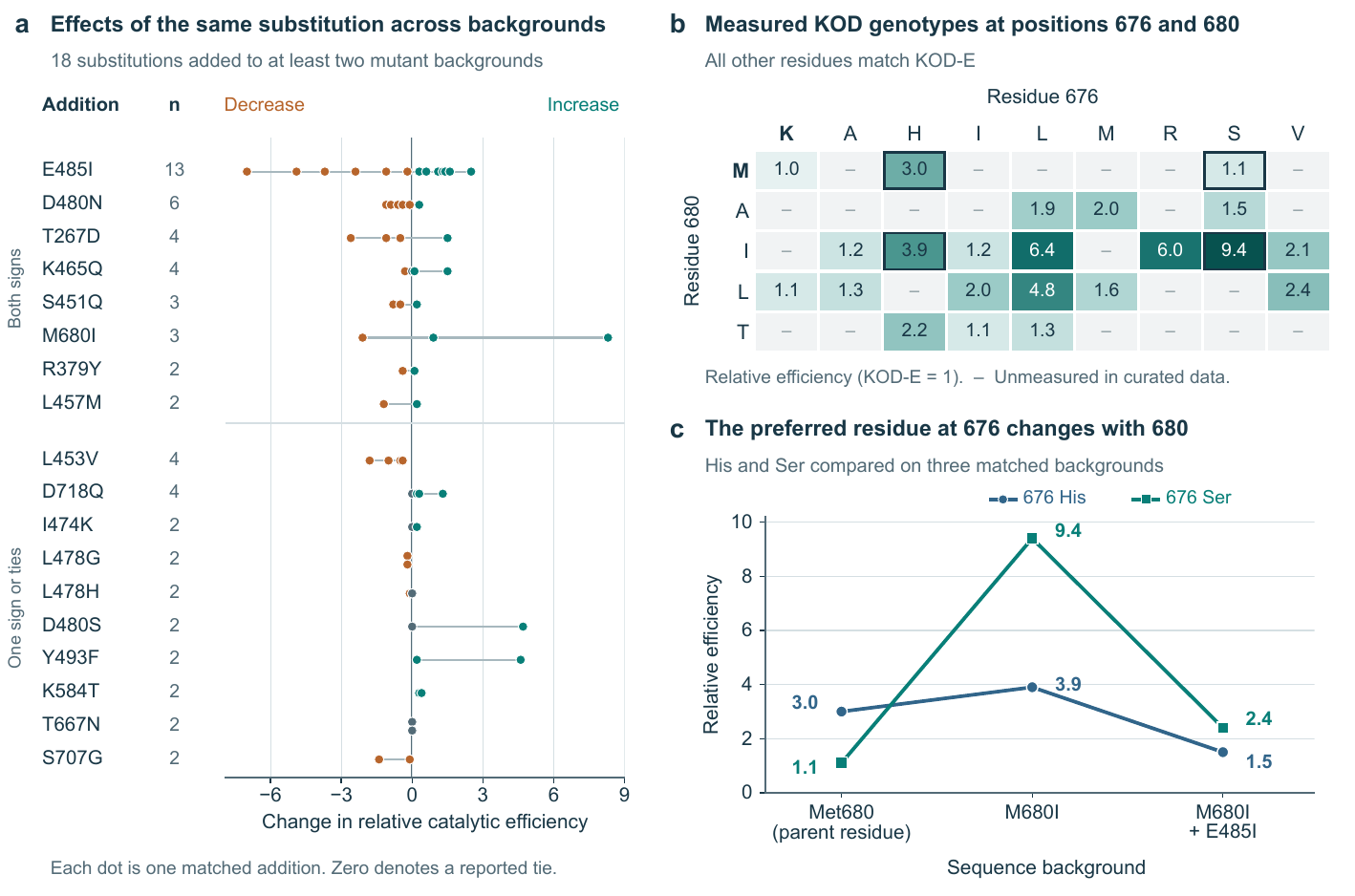}
\caption*{\textbf{Fig. 7 $|$ Benefit of additional substitutions across sequence backgrounds, and reversals of rank between positions, for KOD-E.} \textbf{a}, The 18 additional substitutions that occur in at least two mutant backgrounds, with one point per background-matched addition; the horizontal axis is the change in relative catalytic efficiency and the upper part holds the 8 substitutions that occur in both directions. \textbf{b}, Residue combinations when only positions 676 and 680 are varied; cells give the efficiency relative to KOD-E and a dash marks a combination not assayed in these data. \textbf{c}, His compared with Ser in three matched backgrounds: His leads in the parental Met680 background, Ser overtakes it in Ile680, and the gap narrows when E485I is added.}
\end{figure}

\begin{figure}[!htbp]
\centering
\includegraphics[width=0.86\linewidth,height=0.62\textheight,keepaspectratio]{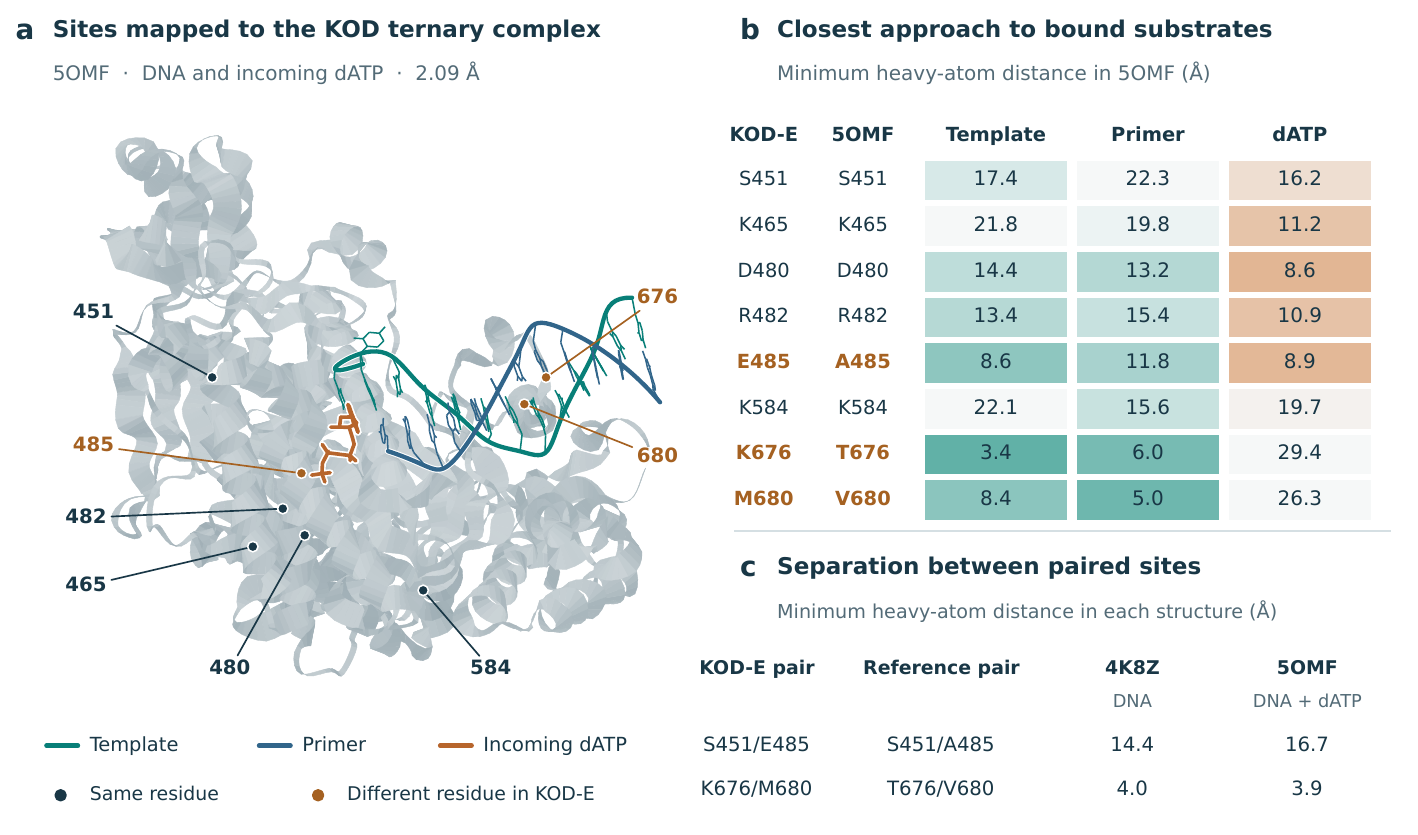}
\caption*{\textbf{Fig. 8 $|$ Location of two position pairs on the KOD ternary complex and their shortest distances to the substrate.} \textbf{a}, Positions mapped onto 5OMF (resolution 2.09 \AA), with residues identical to KOD-E in dark and residues that differ in KOD-E in orange. \textbf{b}, Shortest heavy-atom distance from each position to the template strand, the primer strand and the incoming dATP. \textbf{c}, Shortest distance between the positions of each pair, for both pairs, in 4K8Z and in 5OMF. All distances were computed by EnzymeRSI without access to construct activity values.}
\end{figure}

\FloatBarrier
\phantomsection
\section{3 Discussion}\label{ux8ba8ux8bba}

Handing the question of what to measure next to an agent first exposes not predictive accuracy but the separability of pool construction from ranking. The pool EnzymeRSI froze before the RLuc measurements had a high hit rate, yet design rank within it was almost uncorrelated with measured rank. An agent improving against a primary metric may therefore keep gaining on that metric while delivering nothing at the bench. Active learning and protein-language-model-driven directed evolution [15--17] optimise which candidates to assay on the premise that the method of judgement is fixed; this separation suggests that premise fails in enzyme engineering.

The problem thus shifts from building a more accurate scorer to asking whether decision rules can be recompiled against the objective. In A3A, moving only the cap on unmodified-C loss switched the top-ranked variant; in KOD-E, two objective functions gave different next steps from the same parent and moved in opposite directions at the same triple mutant. An optimum that jumps with the boundary condition cannot be fixed in advance inside a static scorer and must be recompiled at each run, whereas the ranking gap exposed by RLuc lies at the method layer and is a trigger condition for the inner loop. Using no activity values, the system also separated close-contact from distant position pairs on structure alone.

The regularities obtained once the harness is itself treated as an object of evaluation extend beyond these three enzymes. Existing harness self-improvement work takes the pass rate and cost of coding benchmarks as its endpoint [36]. Here, every attempt made by one framework consisted of only three linear transformations, which scored identically to the starting point because the Pearson correlation is invariant to a positive affine transform. Existing gates are built mostly on failure paths, and EnzymeRSI accordingly moved detection of repeated predictions and invariant transforms to the host side. We further observed the host-provided cross-validation split being reused 20 times while its scoring call was invoked twice, suggesting that signals meant for an autonomous researcher should ride on actions that the researcher must execute anyway.

Three boundaries apply to what has been shown here. First, the outer-loop gain of 0.044 is measured on the development set, the set the harness was permitted to use during evolution; of the five endpoints ranked against the official record, only the CD19 first place was produced by the outer loop itself, while the expression first place derives from the tool combination already present at the starting point. The two are consistent with the separation this paper proposes: the CD19 gain came from a change to the variant input at round 12, that is, from what the harness reads rather than from how candidates are ranked, whereas the RLuc gap is a ranking failure that the harness did not repair. Second, the ranking gap exposed on RLuc is named here as a trigger condition for the inner loop but has not yet been closed: the inner loop was exercised on a stability task and an extrapolation task, not against this gap, so whether method development can repair within-pool ranking remains open. Third, the three enzyme tests examine the system's experimental decision-making rather than a complete dry-to-wet cycle driven by the two loops; the commission, execution and receipt path of Section 4.6 is specified but has not yet carried a multi-round campaign end to end, which we treat as a scope decision for this report rather than an unresolved capability.

Four steps follow directly. (1) Trigger the inner loop against the RLuc ranking gap and report whether the correlation between design rank and measured rank improves under the same nomination budget. (2) Run one campaign in which the two loops commission, receive and respond to several rounds of physical measurement, so that the interaction term of Section 4.9 is estimated on wet-lab rather than dry-lab endpoints. (3) Replicate the development-task comparisons across independent evolution seeds, so that framework-level statements rest on the paired statistics specified in Section 4.9. (4) Release the host-side gates, namely the repeated-prediction and invariant-transform pre-checks together with the sealing mechanism of Section 4.10, as a reusable evaluation contract, since the regularities they expose are not specific to enzyme engineering.

Scoring the organisation of research and the development of methods as two separate objects is what made these boundaries visible in the first place; it is also the standard against which the next round of work should be measured.

\phantomsection
\section{4 Methods}\label{ux65b9ux6cd5}

\phantomsection
\subsection{4.1 Task representation and shared research state}\label{ux4efbux52a1ux8868ux793aux4e0eux5171ux4eabux7814ux7a76ux72b6ux6001}

A task is defined as \(\text{q}=\left( \text{x}_{\text{0}},\text{g},\text{c},\text{D}_{\text{vis}},\text{B} \right)\), where \(\text{x}_{\text{0}}\) is the parent sequence, \(\text{g}\) comprises the performance objective and the constraints that may not be violated, \(\text{c}\) covers constructs, substrates, reaction conditions, endpoints and references, \(\text{D}_{\text{vis}}\) is the data the task may legitimately access, and \(\text{B}\) is a multidimensional resource budget. Readouts under different substrates or conditions are retained as separate components and are handled by constrained optimisation, worst-case optimisation or Pareto comparison. At each step the system binds the current harness version \(\text{H}_{\text{k}}\) and algorithm version \(\text{A}_{\text{j}}\):

\begin{equation*}
\text{q}=\left( \text{x}_{\text{0}},\text{g},\text{c},\text{D}_{\text{vis}},\text{B} \right),\quad\quad\text{a}_{\text{t}}\sim \text{π}_{\text{H}_{\text{k}}}\left( \,\cdot\mid\text{s}_{\text{t}},\text{A}_{\text{j}},\text{q} \right)
\end{equation*}

The research state \(\text{s}_{\text{t}}\) records the current candidates, the actions already taken, the available evidence, the trajectory and the remaining resources. Evidence \(\text{E}\) records what was observed, under what conditions, from what source and with what uncertainty, and computational predictions, measurements from the literature, historical experiments and new experiments of the current round carry different source labels. The research trajectory \(\text{T}\) links the evidence visible at the time, the hypotheses to be distinguished, the expected observation, the action chosen, the result returned and the subsequent decision, and labels each conclusion as supported, refuted, not distinguished or not assessable. Evaluation sets, test labels and budget boundaries are managed by an independent evaluation layer and are not modified by any candidate version.

Within the scientific constraints and the budget, the system maximises task quality, and outer-loop development, inner-loop training and failed retries are all charged to cost:

\begin{equation*}
\underset{\text{π},\text{H},\text{A}}{\text{max}}\text{\;}\text{E}\left\lbrack \text{Q}\left( \text{τ} \right) \right\rbrack\quad\quad \text{s.t.}\quad \text{C}\left( \text{τ} \right)\preceq \text{B},\quad \text{τ}\in \text{F}\left( \text{q} \right)
\end{equation*}

\phantomsection
\subsection{4.2 Outer loop: versioned evolution of the harness}\label{ux5916ux73afharness-ux7684ux7248ux672cux5316ux6f14ux5316}

From the trajectories of development tasks the outer loop identifies failure modes, which include incomplete parsing of the objective, a mismatched choice of tool, an inappropriate order of steps, evidence that the next step does not consume, and budget spent on ineffective repetition. For one failure mode it proposes one explicit hypothesis and one minimal consistent change bundle. The bundle records the parent version, the difference, the metric expected to improve, the falsification condition, the range of tasks to which it applies and the evaluation plan, which prevents post hoc explanation from replacing prior prediction; every item in a bundle can be disabled individually so that items can be ablated one at a time. The candidate version and the current version are compared in pairs under the same foundation model, tool permissions, visible data and budget. The outer loop may access development evidence only; the validation set is used solely for the stopping and version-selection rules agreed in advance, and the final test set takes no part in generating changes.

\begin{equation*}
\widetilde{\text{H}}=\text{G}_{\text{H}}\left( \text{H}_{\text{k}},\text{E}_{\text{dev}},\text{T}_{\text{dev}} \right),\quad \text{H}_{\text{k}\text{+1}}=\left\{ \begin{matrix}
\widetilde{\text{H}} & \text{if\:accepted} \\
\text{H}_{\text{k}} & \text{otherwise}
\end{matrix} \right.\ 
\end{equation*}

A version may replace the current version only when its products are valid, the primary metric meets the adoption condition set in advance, the key constraints are satisfied and the cost is acceptable. A candidate whose overall gain is insufficient but which gives a reproducible gain within a predefined task category is retained as a category-specific version. A task already in progress stays bound to the assets with which it started. An adoption event must update the version register and record which version the next run actually consumed: generating a candidate, or writing out a recommendation to adopt, does not by itself close the loop.

The empirical evaluation of automated evolution of \(\text{G}_{\text{H}}\) is given in Section 2.3.

\phantomsection
\subsection{4.3 Inner loop: recursive improvement of algorithmic methods}\label{ux5185ux73afux7b97ux6cd5ux65b9ux6cd5ux7684ux9012ux5f52ux6539ux8fdb}

The inner loop is concerned mainly with proposing and comparing new methods. It is invoked only when the method itself has become the bottleneck, and only after the other possible causes have been excluded in turn; this section describes its design protocol. When a task enters the system, the agent first attempts a solution within the current harness version and algorithm version. Missing packages, permission problems and input-format errors are engineering faults and are repaired directly. Reordering steps, switching to an already approved tool and choosing to re-measure are adaptations within the task. Workflow failures such as incomplete parsing of the objective, evidence that the next step does not consume, and budget occupied by ineffective repetition are handed to the outer loop as a minimal consistent change bundle (Section 4.2). The inner loop is triggered only when the options at the tool and workflow layers have been exhausted and a specific gap in method capability can be named. Such a gap may take the form of insufficient ranking capability, failure to extrapolate beyond the observed conditions, distorted confidence, limited search coverage, or a query gain that falls below the noise scale several times in succession, where the noise scale is estimated from baseline replicates and from resampling grouped by protein. A valid measurement returned from the World Loop that systematically contradicts the predicted ranking or the stated uncertainty is also grounds for triggering, whereas missing measurements and instrument faults enter execution diagnosis. On triggering, the agent submits a structured method development request that states the current method version, the failure mode, the data that may be used, the target metric and an independent resource ceiling; the cost of development is charged to the total task budget.

\begin{equation*}
\widetilde{\text{A}}=\text{G}_{\text{A}}\left( \text{A}_{\text{j}},\text{d}_{\text{t}},\text{D}^{\text{allowed}},\text{B}_{\text{A}} \right),\quad \text{A}_{\text{j}\text{+1}}=\left\{ \begin{matrix}
\widetilde{\text{A}} & \text{if\:accepted} \\
\text{A}_{\text{j}} & \text{otherwise}
\end{matrix} \right.\ 
\end{equation*}

Once accepted, development proceeds asynchronously in an isolated environment. The agent decides the scientific route and selects models and methods itself, over a search space that covers sequence or structure representation, reweighting of training samples, model architecture, loss, fine-tuning and distillation, ensembling and inference calibration, together with mutation operators, recombination schemes, acquisition functions and batch selection rules. The host does not guide the method. It verifies identity, budget, permissions and state, and funnels consequential actions such as training execution and evaluation through its own entry points; the sealed test set is not visible to the agent. Every formal submission is attached to a falsifiable hypothesis that states the expected observation and the condition that would refute it, and the parameters to be changed and the data to be used are declared before the run. The host records independently what actually happened, and any discrepancy is marked as unverified and affects subsequent credence and parent selection. When the gain stays below the noise scale, or the same class of failure recurs, the host issues a non-binding recommendation to change route. A separate supervising agent holds the complete research trajectory, but only observes and reports; it does not alter budgets or intercept submissions.

Feedback operates at two levels. Inner evaluation, which does not consume the formal quota, runs on host-fixed folds that respect the biological holdout structure and returns per-protein error together with the paired difference against the current best, so that search takes place mainly at the inner level and formal evaluation serves only to confirm. Formal evaluation is scored by an independent evaluation layer: an execution failure is not recorded as a score; a prediction identical to an earlier submission is returned before scoring and consumes no quota; and a candidate that differs from its parent or from the starting point by an affine transform alone is marked as giving no gain in ranking (Section 2.2 and Table 4). Credence is granted at three levels, namely that the score is valid, that the method claim holds, and that the scientific conclusion holds; none of the three implies another. A MethodBundle produced by development contains code or configuration, a dependency lock, weights or a training recipe, the data and evaluation versions, the range of applicability and the known defects. Before delivery it must be reloaded in a new process and a new sandbox and must perform genuine inference on shuffled inputs and on unseen mutations, in order to verify that it is not reading a cache or falling back to the output of the starting point. A returned MethodBundle is registered as a candidate only; evaluation and adoption are carried out by the independent evaluation layer and the adoption module above it, and the research agent may not modify the evaluation rules or declare adoption itself.

Evaluation uses the protein, or a group of related proteins, as the statistical unit, and the primary endpoint is the group-level paired gain. Three controls at equal budget are run: active learning with a fixed recipe, continued training, and search restarted from scratch, so that a persistent change of method can be distinguished from a larger number of attempts (Section 4.9). A fall in the training loss, or a rise in a single development-set score, is not grounds for adoption. The adoption rule is agreed before the comparison and yields one of global adoption, scoped retention or rollback. An adoption must be written to the version register and confirmed by a consumption receipt returned by the caller, which states whether loading succeeded and on which inputs the version ran; neither a saved artefact nor a recommendation to adopt counts. Absence of improvement, exhaustion of the budget and failure to load are reported as such and are not counted as a successful method delivery. The joint gain of an algorithm version and a harness version is assessed separately through the interaction term of a four-cell paired comparison (Section 4.9).

\phantomsection
\subsection{4.4 Candidate generation, acquisition and batch selection}\label{ux5019ux9009ux751fux6210ux91c7ux96c6ux4e0eux6279ux6b21ux9009ux6837}

The candidate generator may call on position-wise substitution, combinatorial mutation, homology information, protein language models or structure-constrained methods, but every candidate must specify its parent, its complete sequence and its set of substitutions. Selection combines predicted performance, uncertainty, novelty relative to the samples already measured, and execution cost. The acquisition function template is given below; its specific form is a method asset that the inner loop may evolve:

\begin{equation*}
\text{α}_{\text{t}}\left( \text{x} \right)={\widehat{\text{μ}}}_{\text{t}}\left( \text{x} \right)+\text{β}_{\text{t}}{\widehat{\text{σ}}}_{\text{t}}\left( \text{x} \right)+\text{η}_{\text{t}}\text{\,Novelty}\left( \text{x} \right)-\text{λ}_{\text{t}}\text{\,Cost}\left( \text{x} \right)
\end{equation*}

The normalisation and the weight of each component are fixed during development and then frozen. \(\text{σ}\) is used as a probability only when the uncertainty has passed a coverage or error-stratification check, and otherwise serves only as a model-disagreement term. Batch selection balances acquisition value against batch diversity within the capacity and the budget, and reserves positions for references, replicates and quality control. The acquisition score determines what to measure and the actual measurement determines how well that candidate performs; the two are stored separately.

\phantomsection
\subsection{4.5 Operon-S1 fast decisions and ProtoPilot orchestration}\label{operon-s1-ux5febux51b3ux7b56ux4e0e-protopilot-ux7f16ux6392}

EnzymeRSI can complete the evolution on its own. ProtoPilot [41] is responsible mainly for wet-lab orchestration: it develops the protocol and the standard operating procedure (SOP) from the research intent, generates device control code, executes after validation, and revises the procedure in the light of wet-lab feedback. High-frequency, well-defined experimental decisions are delegated by ProtoPilot to Operon-S1, and the two together accelerate decision-making. Operon-S1 is a locally deployed small model for fast experimental decisions, designed for low cost and low latency, for keeping data on site, and for iterative adaptation as experimental feedback accumulates, so that the large-model agent can concentrate on reasoning, exploration, experiment execution and feedback analysis. The routing rule is that decisions over a limited action space, such as candidate pre-screening, batch admission and QC judgement, go to Operon-S1, whereas hypothesis generation, diagnosis of method bottlenecks and the design of change bundles remain with the agent. 

\phantomsection
\subsection{4.6 World Loop: cloud laboratory commissions and receipts}\label{world-loopux4e91ux5b9eux9a8cux5ba4ux59d4ux6258ux4e0eux56deux6267}

EnzymeRSI outputs the objective, the candidates, the sample requirements, the measurement endpoints, the references, the replicate design and the budget. Wet-lab orchestration is carried out by ProtoPilot, which converts these requirements into a protocol, an SOP and device control code, checks them against materials and instrument capability, and submits them for execution once validated [41]. Scheduling, running, measurement and return are handled by a cloud-to-edge laboratory execution system that provides cloud scheduling, edge orchestration, hardware abstraction and an execution receipt mechanism [42]. The same task contract can address a single instrument, a workcell, a production line or a cloud laboratory, with an execution adaptation layer absorbing the differences between devices.

Experiments, samples, containers and steps carry unique identifiers. Commissions, actions and receipts are linked by an execution token, and the planned and the actual execution parameters are stored separately. Sample barcodes link lineage, batch, concentration, location and storage history, and raw files are attached to the evidence record by file identifier and checksum. Execution receipts and actual measurements are stored apart: a step that returns success evidences a software event only, whereas the quantity actually transferred, the reaction conditions and the identity of the product still require the corresponding measurement and quality-control support. A sample that passes QC but shows low activity enters learning as valid biological evidence, whereas missing measurements, invalid measurements and instrument faults follow the execution diagnosis path. Each new measurement first completes identity matching, verification of conditions, normalisation to the reference and a quality judgement before \(\text{E}\) is updated, and the record states whether that result changed the candidate ranking, the algorithm diagnosis or a harness hypothesis. Historical replay, measurements from the literature and physical execution in the current round retain different labels, because they answer different verification questions.

\phantomsection
\subsection{4.7 Tiered scoring and version adoption}\label{ux5206ux5c42ux8bc4ux5206ux4e0eux7248ux672cux91c7ux7528}

The scoring layer reports the protein, algorithm and system tiers separately, each with its reference, direction, sample unit, evaluation set and uncertainty. Failures of candidate generation, failures of execution and results that cannot be assessed are reported separately and are not filled in as biological zeros (Table 6). The adoption rule is fixed before the comparison and yields one of three outcomes: global adoption, scoped retention or rollback.

\phantomsection
\subsection{4.8 Empirical evaluation setup}\label{ux5b9eux8bc1ux8bc4ux4ef7ux8bbeux7f6e}

The two development tasks use the same evaluation contract. The ThermoMPNN stability task uses one training set (215,731 records, 239 proteins) and one development set (27,328 records, 31 proteins), with the official ThermoMPNN predictor as the common starting point, whose development-set PCC is 0.804358. The TadA task uses the amino-acid view development task of TadA-Bench, with the development-set Spearman correlation as the primary metric. Each run is given 20 formal evaluations, a research wall-clock limit of 48 h and a memory limit of 96 GiB. A formal evaluation token consumes quota as soon as it is submitted, and a submission with invalid predictions is counted in the same way. Each evaluation is scored independently by the host scorer and archived with a signature.

The research base is a locked version of the general-purpose coding agent QwenCode (v0.26.5), run in a native headless session. The distribution version, backend model identifier, prompts, sampling configuration, tool definitions and session continuation rules are all frozen as run configuration, and context compaction, retry and recovery retain their native behaviour. The host provides only the task description, the submission command and independent scoring, and does not intervene in research decisions. On top of this base, EnzymeRSI adds five host-side research support layers (Table 4) in two classes, execution enhancement and cross-framework integration. The first allows long training runs to proceed without a foreground timeout limit and replaces opaque transport errors with diagnosable exceptions. The second converts into host mechanisms those steps that proved effective, or were found missing, across runs of multiple frameworks: a research state query that returns the current best and the full submission history from the signed record; a repeated-prediction pre-check that sends back any submission whose prediction vector is identical to an earlier one; and a delivery-manifest pre-check that verifies the submission list against the packaging rules. A return from either pre-check consumes no evaluation quota.

Ten open-source or internal research frameworks were run under the same task directory, training and development split, official starting point and host scorer: OpenRSI, AReaL-RSI, AgenticEvolve, ShinkaEvolve, MLEvolve, RD-Agent, OpenEvolve, TusoAI, AI-Scientist-v3 and ERA. Each framework was run for its own native number of outer-loop rounds (Table 2), failed evaluations are counted in the same way, and one such round may produce several scored submissions, so the number of scoring records can exceed the number of rounds. The reconstruction of evaluation-by-evaluation scores and the plotting rules are given in the Appendix.

\textbf{The evaluation contract of the outer loop.} The outer loop was run on one assay from each of five classes of human protein measurement, for 196 rounds in total. The primary metric was spearman\_macro with a minimum acceptable increment of 0.001. In each round the control version and the candidate version were compared in pairs on the same assay, for 142 comparisons in total.

\textbf{The procedure of each round.} Each round runs proposal, application, paired comparison under matched conditions, gating, and adoption or rollback, in that order. The assets that may be changed are limited to skills, MCP tool access, subagents and the system prompt; model weights and the scoring algorithm are unchanged throughout. Any change that fails the gate is rolled back to the parent commit, and the changed files and the commit identifier of every round are retained for traceability. The five endpoints were ranked alongside the 97 published models in the official per-assay ProteinGym record, taken from a frozen snapshot. To confirm that our endpoints and the official record were computed on the same basis, we recomputed the official Average Spearman with the same script and matched the official overall table item by item to three decimal places.

\phantomsection
\subsection{4.9 Evaluation protocol and statistics}\label{ux8bc4ux6d4bux534fux8baeux4e0eux7edfux8ba1}

The development set may be used with its labels to develop the harness and the methods; the validation set is used only for stopping and version selection; and the test set has its labels isolated and network access disabled before final submission, with column-level stripping where model scores and labels share a file. Baselines are chosen and frozen on development or validation information and are not selected by test performance. The primary endpoint is the group-level paired gain:

\begin{equation*}
\text{Δ}\text{Q}=\frac{\text{1}}{\left| \text{G}_{\text{test}} \right|}\sum_{\text{g}\in \text{G}_{\text{test}}}^{\text{\!}}\left\lbrack {\overline{\text{ρ}}}_{\text{g}}\left( \text{candidate} \right)-{\overline{\text{ρ}}}_{\text{g}}\left( \text{reference} \right) \right\rbrack
\end{equation*}

\(\text{ρ}\) is computed per assay, aggregated within a related group, and then compared with equal weight across groups. The primary reference is best\_single\_dev, a single method chosen during development, and a further control is the same agent with evolution disabled. We report the group-level paired difference, a group-level sign-flip test and a 10,000-sample paired bootstrap confidence interval, over at least 3 independent evolution seeds. A single run is not sufficient to support a framework-level ranking, which requires replication across several evolution seeds.

Cooperation between the two loops is tested through the interaction term of a four-cell paired evaluation. All four cells must be executable and semantically consistent; when a method cannot run without its matching harness, the interaction must not be computed by filling in a zero:

\begin{equation*}
\text{I}=\text{Y}\left( \text{H}_{\text{1}},\text{A}_{\text{1}} \right)-\text{Y}\left( \text{H}_{\text{1}},\text{A}_{\text{0}} \right)-\text{Y}\left( \text{H}_{\text{0}},\text{A}_{\text{1}} \right)+\text{Y}\left( \text{H}_{\text{0}},\text{A}_{\text{0}} \right)
\end{equation*}

Strategy discovery in the inner loop is tested by dense landscape replay, which measures the best observed performance, the query efficiency and the area under the curve at equal query budget. Its transfer to physical experiments is not reported here; the three enzyme cases of Section 2.4 test the decision rules themselves, not a strategy produced by the inner loop. Equal-budget comparison includes active learning with a fixed recipe, continued training at equal budget, and search restarted from scratch, so that gains from persistent accumulation can be distinguished from gains from a larger number of attempts. The number of instrument events does not serve as an independent biological sample size.

\phantomsection
\subsection{4.10 Tiered evaluation resources and the sealing mechanism}\label{ux5206ux5c42ux8bc4ux6d4bux8d44ux6e90ux4e0eux5c01ux5b58ux673aux5236}

The evaluation reported here does not rest on a single benchmark. Four tiers of resource carry different conclusions (Table 7). The static task tier (\texttt{evolution-\allowbreak{}v1}) performs per-assay self-improvement with related groups as the unit of control, and carries the five outer-loop endpoints and their ranking alongside the official record. The dense replay tier (\texttt{evolution-\allowbreak{}replay-\allowbreak{}v1}) fixes the query budget and uses a dense landscape as an in silico oracle, and carries the comparison of sequential decision-making and budget efficiency; its 133 environments are split into 88 development, 5 validation and 40 test. The real-time extrapolation tier (\texttt{TadA-Bench}) preserves the temporal order of 31 rounds of PANCE evolution and carries the test of whether a system can learn from previous experiments how to conduct the next one. The external transfer tier carries transfer across proteins and data sources, and supplementary evidence. The boundary of each tier is stated in the right-hand column of Table 7, and the tiers may not borrow from one another.

\textbf{Sealing is more than a declaration.} Isolation on the test side is implemented by the following mechanisms. The 40 test environments of \texttt{evolution-\allowbreak{}replay-\allowbreak{}v1} are refused directly by the runner, and only an explicit \texttt{-\/-final} admits them, with an audit record. The parquet files and tables corresponding to rounds 29--31 of \texttt{TadA-Bench}, the final round of the three closed-loop campaigns and the final PANCE cycle are sealed with permission 000. A strategy may obtain revealed scores and public zero-shot columns only through \texttt{ArmContext}, and opening an assay CSV or an oracle table directly invalidates the run. The sha256 of every split is frozen in its own \texttt{manifest.json}, and a change requires a new suite identifier rather than a re-split in place. The statistical unit is always the related group rather than the assay or the variant, and paired tests are performed group by group.

The composition of the replay records and the details of the query protocol are given in the Appendix.

\phantomsection
\subsection{4.11 Wet-lab work and structural analysis for the three enzymes}\label{ux4e09ux9176ux6e7fux5b9eux9a8cux4e0eux7ed3ux6784ux5206ux6790}

The three-enzyme work reported in this section was carried out by the experimental team, with EnzymeRSI supplying the candidate nominations and the selection rules. The three enzyme studies were executed in a cloud-laboratory environment through an integrated protocol-orchestration and fast-decision workflow. Experimental intent was compiled into machine-readable procedures with predefined decision checkpoints for construct and sample identity, plate admission, control completeness, assay validity and quality control. At each checkpoint, the current state was evaluated against fixed criteria and mapped to a bounded action, such as proceed, hold, request recheck or exclude. Physical operations and measurements were performed by the cloud-laboratory instruments, with operator intervention for steps outside device capability. Sample identities, execution records, raw files and quality-control outcomes were linked before downstream analysis. These experiments tested single-round decision and execution support; they did not constitute a multi-round campaign in which the recursive-improvement loops autonomously commissioned and adapted successive experimental rounds.

RLuc: mutant construction and luminescence assay

\textbf{Construction of mutant plasmids.} The nomination stage used only single-point measurements from the literature together with sequence and structural information, and covered 40 single substitutions at 31 positions. The design record was frozen before the wet-lab measurements. The wet-lab work used \emph{Renilla} luciferase (UniProt P27652, 311 aa) as the parent. The coding sequence was codon-optimised, synthesised by Azenta, cloned into the pET-21b(+) expression vector and transformed into \emph{Escherichia coli} BL21(DE3). For each nominated amino-acid substitution, mutagenic primers were designed with Vazyme CE Design and synthesised by Sangon. Site-directed mutagenesis was performed in 50 µL reactions containing 1 U Phanta Flash high-fidelity DNA polymerase (Vazyme), parental plasmid template at a final concentration of 0.2 ng/µL and 0.4 µM mutagenic primers. After amplification, 1 µL DpnI was added per 50 µL of PCR product and digested at 37 °C for 30 min. Digested products were chemically transformed into BL21(DE3) and single colonies were picked for expression screening. Construct designs and measurement wells were matched one to one by construct number and amino-acid substitution, and the designed sequences were aligned against the parent to confirm the identity of each single substitution.

Within the RLuc workflow, construct identity, plate admission, WT and blank control completeness, signal-window completeness and replicate quality were checked before the corresponding readouts were admitted to analysis. Physical screening was performed using the instruments and conditions described below.

\textbf{High-throughput colony culture and preparation of crude enzyme extract.} Single colonies were picked with a QPix automated colony picker and inoculated into 96-well plates containing 150 µL LB medium with 100 µg/mL ampicillin, then grown overnight at 37 °C and 200 rpm in a Cytomat 2 automated incubator. The following day, cultures were transferred to fresh LB medium containing ampicillin at the same concentration and grown at the same temperature and shaking speed to an OD600 of 0.6--0.8. IPTG was added to a final concentration of 0.5 mM and expression was induced for 16 h at 16 °C and 200 rpm. Then 3 µL of 10$\times$ BugBuster lysis reagent was added per 150 µL of culture, mixed, and left at 4 °C for 30 min; the resulting crude lysate was used directly for the chemiluminescence assay.

\textbf{Kinetic initial screen, whole-library measurement and calculation of integrated luminescence.} Crude enzyme extract (20 µL) was transferred to the reaction well, and a Tecan Infinite 200 PRO microplate reader injected 100 µL of substrate buffer containing 0.5 µM native coelenterazine (MGI) automatically, for a total reaction volume of 120 µL. After injection the plate was shaken orbitally for 1 s and signal was collected continuously for 300 s in chemiluminescence mode at 28 ± 2 °C, mostly at 0.3 s intervals. The peak of each well in the initial screen was defined as the maximum value observed within the first 120 s, normalised to the arithmetic mean of the wild-type (WT) reference-well peaks of the same batch; plates 2 and 3 of the first batch shared WT references. The whole-library single-point measurement was performed on three plates, with each candidate variant cultured, expressed and lysed independently in triplicate. Each candidate was therefore measured once, in three independent preparations. and with three WT reference wells and three blank medium control wells per plate. The integrated readout was defined as the area under the luminescence curve (AUC, that is, the integrated luminescence) of an equal volume of crude enzyme extract within a selected window over the first 290.2 s, and the relative integrated luminescence is the ratio of the integral of each sample to the arithmetic mean of the WT integrals on the same plate (Fig. 4). Of the 120 independent candidate preparations (40 candidates $\times$ 3 replicates), 111 entered the main analysis, 8 further quantifiable samples were included in the sensitivity analysis, and the remaining 1 was excluded because the measurement did not cover the complete integration window. For the robustness check on these three candidates, the denominator and the relative fold were recomputed after removing each WT reference well on the same plate in turn, in order to quantify the effect of well-to-well variation on the threshold judgement.

APOBEC3A (A3A): construction and expression, dual-substrate sequencing and structural localisation

For A3A, the workflow linked each construct to its expression, purification, paired CG and CT reactions, library and sequencing records. Paired-measurement completeness, reaction and library quality, and eligibility for dual-context ranking were checked before conversion rates were calculated.

\textbf{Construction of the expression vector, site-directed mutagenesis and protein purification.} A synthetic wild-type \emph{APOBEC3A} coding sequence was used as the parent and first cloned into a His-tagged pET-28a vector. The A3A-WT-His fragment and the pET-41b(+) vector backbone were amplified separately with 2$\times$ Phanta Flash Master Mix (Vazyme, P510-01) and joined by homologous recombination with the CloneExpress II One Step Cloning Kit (Vazyme, C112-01), giving the pET-41b(+)-A3A-WT expression plasmid with both GST and His tags. Mutagenic primers were designed for each amino-acid substitution, and variant plasmids were amplified from this plasmid, assembled, transformed into \emph{E.~coli} BL21(DE3) and grown on kanamycin LB plates at 37 °C for 12--16 h. Two to three single colonies per construct were verified by sequencing. Verified single colonies were inoculated into kanamycin LB in well plates and grown at 37 °C for 12--16 h, then transferred in proportion to fresh antibiotic-containing LB for further plate culture. At an OD600 of 0.6--0.8, IPTG was added to a final concentration of 1 mM and expression was induced at 20 °C for 12--16 h. Products were purified in-plate by Ni-NTA agarose affinity chromatography, and the purified protein was used for the deaminase activity assay.

\textbf{Activity screen, dual-substrate reaction, library preparation and high-throughput sequencing.} Qualitative screens in the CG and CT contexts used the corresponding single-stranded oligonucleotide substrates. Each 20 µL reaction contained 1 µg purified A3A protein, 0.2 µL of 100 µM oligonucleotide and 18.8 µL of 1$\times$ A3A Reaction Buffer (Abclonal, RM20819) and was incubated at 37 °C for 3 h; 10 µL of a mixture containing 1 U USER enzyme was then added and incubated at 37 °C for a further 1 h. A 10 µL aliquot of the product was resolved on a 20\% urea--polyacrylamide gel (urea-PAGE), stained and imaged, and deaminase activity was confirmed from the cleavage band of the uracil-containing product. The quantitative dual-substrate assay used an unmodified and a methylated oligonucleotide of identical base sequence (Sangon; the latter carrying 5mC at the three specified target C positions), reacted independently. Each 100 µL reaction contained 1 µL enzyme solution (1 µg A3A protein), 20 µL of the specified oligonucleotide (200 ng), 10 µL APOBEC Reaction Buffer (Abclonal, RM20819), 1 µL of 20 mg/mL recombinant albumin (Abclonal, RM20543) and 68 µL nuclease-free water. With the thermal cycler lid at 50 °C, reactions were incubated at 37 °C for 3 h, after which 200 µL of purification beads (Vazyme, N411-01) were added to purify the product. Two rounds of library amplification were then performed with Gloria U 2$\times$ HS Master Mix (Abclonal, RK20722), both with the lid at 105 °C. The first PCR (98 °C for 3 min; 10 cycles of 98 °C for 30 s, 60 °C for 30 s and 72 °C for 30 s; 72 °C for 5 min) was purified with 50 µL beads. The second PCR introduced phosphorylated ends and sample barcodes, with the annealing temperature changed to 56 °C and the cycle number to 8 and all other conditions as in the first round, and the product was again purified with 50 µL beads. Of the second-round product, 200 ng was circularised with the MGIEasy circularisation kit (MGI, 1000005259), and 10 ng of the circularised library was used to prepare DNA nanoballs (DNBs) for single-end sequencing (effective read length 75 nt) on an MGISEQ-2000RS platform with the FCL SE100 kit (MGI, 940-001805-00). The conversion rate was defined as the proportion of reads in which a specified target C or 5mC position was read as T. The values reported here are endpoint readings taken after 3 h under the conditions above, and WT enzyme prepared in the same laboratory batch served as the reference against which variants were compared. In total, 78 variants were measured in the CG context, of which 75 were also measured in the CT context and entered the paired dual-context analysis; the three variants measured in CG only were K47V, K137T and C171R.

\textbf{EnzymeRSI dual-context selection rule and structural distance calculation.} At the rule-generation stage, the design objective supplied by the researchers was to cap the C conversion rate in both the CG and CT contexts while maximising the lower of the two 5mC conversion rates. EnzymeRSI formalised this as a constrained filter with a minimax ranking rule: for the 75 variants with a complete set of four paired measurements, the larger of the two C conversion rates, max(C\textsubscript{CG}, C\textsubscript{CT}), served as the hard constraint, and the smaller of the two 5mC conversion rates, min(5mC\textsubscript{CG}, 5mC\textsubscript{CT}), served as the ranking index; within the eligible set satisfying the constraint, the variant with the highest ranking index was retained, ranked at full floating-point precision, which generated Fig. 5a. The substrate preference difference was defined as the 5mC minus C conversion difference within a context, in percentage points. The preference difference of WT is negative in both contexts, and a variant whose preference difference is positive in both the CG and the CT context was classified as showing a reversal of substrate preference in both contexts (Fig. 5b, c). For structural localisation, the A3A parent sequence (199 aa) was aligned to chain A of the complex crystal structure 5KEG (resolution 2.20 \AA{}) [40], with which it is identical apart from the inactivating mutations E72A and C171A introduced for crystallisation. Geometric distances were computed over heavy atoms with positive occupancy: the shortest distance from the Cε2 atom of the Tyr130 aromatic ring to the C5 atom of the target cytosine at position 3 of chain B is 3.664 \AA{} (given as 3.66 \AA{} in the main text), and the distance from its phenolic hydroxyl oxygen (OH) to the OP2 atom of that nucleotide phosphate group is 2.666 \AA{}.

KOD-E: variant preparation, kinetic assay, path pairing and activity-independent structural analysis

For KOD-E, the workflow linked each construct to its parent--descendant identity, control conditions, substrate concentration series and kinetic readouts. Assay validity and completeness of direct comparisons were checked before kinetic fitting and subsequent path analysis.

\textbf{Construction, expression, purification and kinetics of modified-substrate incorporation.} The engineered archaeal polymerase KOD-E (774 aa in full length, with the initiating Met numbered 1; coding sequence 2,325 nt including the stop codon) served as the parent, and the amino-acid substitutions of all variants are numbered according to this parental sequence. The 6$\times$His-tagged coding sequence was cloned into the pD441-pelB expression vector with the Electra Cloning Reagents Kit (DNA2.0, EKT-02), transformed into \emph{E.~coli} BL21 and selected on LB plates containing 50 µg/mL kanamycin; three to five single colonies per construct were verified by sequencing. Verified single colonies were grown overnight at 37 °C in kanamycin LB in well plates, transferred to fresh antibiotic-containing LB plates and grown to an OD600 of 0.8, and IPTG was added to a final concentration of 0.5 mM for overnight induction at 25 °C, with an uninduced parental strain and an empty-vector strain as negative controls. Products were purified in-plate on Ni-NTA agarose. Enzyme activity was assayed with Cy3-labelled modified dATP and an annealed Cy5-labelled double-stranded DNA substrate containing a 25 bp complementary region. Reactions contained 1 µg purified polymerase, 2 µM Cy3-dATP and DNA-Cy5 over a concentration gradient of 0.2--4 µM, in a buffer of 20 mM Tris--HCl (pH 8.5), 10 mM (NH\textsubscript{4})\textsubscript{2}SO\textsubscript{4}, 10 mM KCl and 2 mM MgSO\textsubscript{4}. Reactions were run at 40 °C for 1 h, initial rates were estimated from fluorescence signal collected continuously on a microplate reader, and V\textsubscript{max} and K\textsubscript{m} were fitted with the Michaelis--Menten equation. The relative catalytic efficiency of each variant was defined as the ratio of its k\textsubscript{cat}/K\textsubscript{m} to that of the parent KOD-E, which is set to 1.0. Because the concentration of Cy3-dATP was held constant while that of DNA-Cy5 was varied, these fitted parameters are to be read as apparent kinetic parameters for DNA-Cy5 at a fixed 2 µM Cy3-dATP; the relative measure is not used to report an absolute k\textsubscript{cat}, nor on its own to infer the K\textsubscript{m} for modified dATP.

\textbf{Data cleaning, parent--descendant path pairing and route selection rules.} The KOD-E coding and amino-acid sequences were checked position by position, including Glu at position 458; none of the 414 resolvable substitutions involved that position. Of the 207 construct records, 17 were removed because a readout was missing, the construct definition was ambiguous, the annotated parental residue did not match the parental sequence (KH224 and QH303 differ from the parent at positions 484 and 689 respectively), or readouts for the same construct conflicted, leaving the parent KOD-E and 189 valid variants. Each variant was represented as its complete substitution set relative to KOD-E. A directed parent--descendant edge was created only when the descendant retained every substitution of the parent and added exactly one substitution at a different position, giving 125 single-step comparisons that cover 41 mutant parents and 107 endpoint constructs; one construct may take part in several pairs, and the original numerical precision was retained throughout. For each assayed parent, all direct descendants were enumerated and classified into gain, no-change and loss branches (Fig. 6a, b). For path selection, EnzymeRSI compiled the two classes of engineering objective supplied by the researchers into graph search operators. The first is the highest efficiency among direct descendants, a greedy choice of the assayed single-step descendant with the highest relative efficiency under the current parent. The second is the longest strictly monotonic path of gains, which starts from the parent or a specified origin such as Q461A and extends level by level along assayed edges whose descendant efficiency is strictly higher than that of its direct parent, retains the path reaching the largest number of substitutions, and also marks the descending edge at the point of convergence in the continuous assayed subgraph of Fig. 6c.

\textbf{Background effects, four-construct epistasis controls and structural calculation from activity-free input.} Across the 125 single-step comparisons, additions were grouped by exact amino-acid substitution; a substitution showing both a positive and a negative efficiency difference across sequence backgrounds was recorded as one change in the direction of gain, with a sensitivity check requiring both absolute differences to exceed 0.1 parent units (Fig. 7a). A four-construct control fixes the rest of the sequence, takes two mutant residues at the target position and two assayed states at the background position, and requires measured values for all four genotype combinations; when the difference in rank between the two residues at the target position has opposite signs in the two background states, a reversal of rank is recorded. The 35 directed comparisons reduce to 21 independent four-construct sets after de-duplication by the four original construct record numbers, of which 7 show a reversal of rank; Fig. 7c shows the set with the largest value of the smaller of the two absolute differences, namely His and Ser at position 676 under Met and Ile at position 680. For structural analysis, the mmCIF coordinates of the KOD polymerase binary complex 4K8Z (resolution 2.29 \AA{}) and of the ternary complex with an incoming substrate 5OMF (resolution 2.09 \AA{}) were downloaded, in which chains A, T and P correspond to the polymerase, the template strand and the primer strand, and the incoming dATP in 5OMF is DTP 810. Where required, the A chains of the two structures can be superposed with the TM-align mode of Foldseek, and the difference in residue solvent-accessible surface area before and after DNA binding can be computed with the Shrake--Rupley algorithm using a 1.4 \AA{} solvent probe and 960 or 480 sphere sampling points per atom. Without being given any construct activity value, only the 5OMF and 4K8Z coordinates and the position pairs specified in advance by the researchers, 676/680 and 451/485, were supplied to EnzymeRSI. The program filtered out water, ions and symmetry mates automatically, retained heavy atoms of the first model with positive occupancy in the major conformer (conformation A), verified them against both the polymer sequence and the author residue numbering, and then enumerated the shortest Euclidean distance from every residue, main chain included, to the target chain or the substrate, and reported the closest atom pair.

\phantomsection
\section*{References}\label{ux53c2ux8003ux6587ux732e}
\addcontentsline{toc}{section}{References}

\begingroup\sloppy\small\setlength{\parindent}{0pt}\setlength{\parskip}{2pt}\sloppy\emergencystretch=3em\relax

\begin{enumerate}
\def\labelenumi{\arabic{enumi}.}
\tightlist
\item
 Arnold FH. Directed Evolution: Bringing New Chemistry to Life. \emph{Angewandte Chemie International Edition} 57, 4143--4148 (2018). \url{https://doi.org/10.1002/anie.201708408}
\item
 Yang KK, Wu Z, Arnold FH. Machine-learning-guided directed evolution for protein engineering. \emph{Nature Methods} 16, 687--694 (2019). \url{https://doi.org/10.1038/s41592-019-0496-6}
\item
 Jumper J, Evans R, Pritzel A, et al.~Highly accurate protein structure prediction with AlphaFold. \emph{Nature} 596, 583--589 (2021). \url{https://doi.org/10.1038/s41586-021-03819-2}
\item
 Lin Z, Akin H, Rao R, et al.~Evolutionary-scale prediction of atomic-level protein structure with a language model. \emph{Science} 379, 1123--1130 (2023). \url{https://doi.org/10.1126/science.ade2574}
\item
 Dauparas J, Anishchenko I, Bennett N, et al.~Robust deep learning--based protein sequence design using ProteinMPNN. \emph{Science} 378, 49--56 (2022). \url{https://doi.org/10.1126/science.add2187}
\item
 Biswas S, Khimulya G, Alley EC, Esvelt KM, Church GM. Low-N protein engineering with data-efficient deep learning. \emph{Nature Methods} 18, 389--396 (2021). \url{https://doi.org/10.1038/s41592-021-01100-y}
\item
 Dai F, You S, Zhu Y, et al. Programmable design of functional proteins from natural language (Pinal). \emph{bioRxiv} (2024). \url{https://doi.org/10.1101/2024.08.01.606258}
\item
 Zhang X, Lu WS, Viggiano B, et al. Generative steering to adapt proteins beyond native biological contexts (ProVADA). \emph{bioRxiv} (2025). \url{https://doi.org/10.1101/2025.07.11.664238}. Code: github.com/SUwonglab/ProVADA
\item
 Qian T, Wang W, Gao S, et al. Complex enzyme active site scaffolding by iteratively detuned catalytic guidance (CAGE). \emph{bioRxiv} (2026). \url{https://doi.org/10.64898/2026.10.02.756307}
\item
 Wu Z, Abramson J, Frerix T, et al. Designing enzymes for new-to-nature chemistry and non-natural substrates with AlphaProtein Novo. \emph{bioRxiv} (2026). \url{https://doi.org/10.64898/2026.10.01.756017}. Code: github.com/google-deepmind/alphaprotein-novo
\item
 Cai Q, Xu M, Yan J, et al. Chain of Structures: Connecting Reasoning and Generation for Protein Design. \emph{bioRxiv} (2026). \url{https://doi.org/10.64898/2026.10.09.758133}. Code: github.com/Scinetics-ai/chain-of-structures
\item
 Notin P, Kollasch AW, Ritter D, et al.~ProteinGym: Large-Scale Benchmarks for Protein Fitness Prediction and Design. \emph{Advances in Neural Information Processing Systems 36}, Datasets and Benchmarks Track (2023). \url{https://doi.org/10.52202/075280-2810}
\item
 Tsuboyama K, Dauparas J, Chen J, et al.~Mega-scale experimental analysis of protein folding stability in biology and design. \emph{Nature} 620, 434--444 (2023). \url{https://doi.org/10.1038/s41586-023-06328-6}
\item
 Dieckhaus H, Brocidiacono M, Randolph NZ, Kuhlman B. Transfer learning to leverage larger datasets for improved prediction of protein stability changes. \emph{Proceedings of the National Academy of Sciences} 121, e2314853121 (2024). \url{https://doi.org/10.1073/pnas.2314853121}
\item
 Yang J, Lal RG, Bowden JC, et al.~Active learning-assisted directed evolution. \emph{Nature Communications} 16, 714 (2025). \url{https://doi.org/10.1038/s41467-025-55987-8}
\item
 Jiang K, Yan Z, Di Bernardo M, et al.~Rapid in silico directed evolution by a protein language model with EVOLVEpro. \emph{Science} 387, eadr6006 (2025). \url{https://doi.org/10.1126/science.adr6006}
\item
 Tran VQ, Nemeth M, Bartie LJ, et al. Rapid directed evolution guided by protein language models and epistatic interactions. \emph{Science} 392, eaea1820 (2026). \url{https://doi.org/10.1126/science.aea1820}
\item
 Tan Y, Tian Q, Sun G, et al. A General Harness for Protein Foundation Model Fitness Prediction (VenusREM-Harness). arXiv:2609.34654 (2026). Code: github.com/ai4protein/VenusREM-Harness
\item
 Butcher J, Chen J, Hunt A, et al.~Reliable and Efficient Agentic Protein Design with Toolset Harnessing (Chaperone). \emph{bioRxiv} (2026). \url{https://doi.org/10.64898/2026.10.19.759896}. Code: github.com/RosettaCommons/chaperone
\item
 Isomorphic Labs Team. Accurate Predictions of Novel Biomolecular Interactions with IsoDDE. Zenodo (2026). \url{https://doi.org/10.5281/zenodo.19699685}
\item
 Anew Labs Team, Liu K. AnewDDE: An Agentic Drug Discovery Engine for Biomolecular Interaction Modelling and Closed-Loop Design. \emph{bioRxiv} (2026). \url{https://doi.org/10.64898/2026.09.20.752967}
\item
 Aureka AI OpenDDE project. Folding, Reasoning, and Scaling with Open-source Drug Discovery Engine (OpenDDE). arXiv:2607.03787 (2026). Code: github.com/aurekaresearch/OpenDDE
\item
 Rapp JT, Bremer BJ, Romero PA. Self-driving laboratories to autonomously navigate the protein fitness landscape. \emph{Nature Chemical Engineering} 1, 97--107 (2024). \url{https://doi.org/10.1038/s44286-023-00002-4}
\item
 Burger B, Maffettone PM, Gusev VV, et al.~A mobile robotic chemist. \emph{Nature} 583, 237--241 (2020). \url{https://doi.org/10.1038/s41586-020-2442-2}
\item
 Boiko DA, MacKnight R, Kline B, Gomes G. Autonomous chemical research with large language models. \emph{Nature} 624, 570--578 (2023). \url{https://doi.org/10.1038/s41586-023-06792-0}
\item
 M. Bran A, Cox S, Schilter O, et al.~Augmenting large language models with chemistry tools. \emph{Nature Machine Intelligence} 6, 525--535 (2024). \url{https://doi.org/10.1038/s42256-024-00832-8}
\item
 Latent Labs Team. Latent-Y: A Lab-Validated Autonomous Agent for De Novo Drug Design. arXiv:2603.29727 (2026).
\item
 Muthupandiyan S, Dur TH, Walker M, et al. ReproBAIT: Can coding agents build their own biological AI tools? arXiv:2610.19810 (2026).
\item
 Kim JW, Myaskovsky A, Donosa O, et al. Agent-guided epitope excision enables economical de novo binder design. arXiv:2610.33195 (2026).
\item
 Romera-Paredes B, Barekatain M, Novikov A, et al.~Mathematical discoveries from program search with large language models. \emph{Nature} 625, 468--475 (2024). \url{https://doi.org/10.1038/s41586-023-06924-6}
\item
 Lu C, Lu C, Lange RT, Foerster J, Clune J, Ha D. The AI Scientist: Towards Fully Automated Open-Ended Scientific Discovery. arXiv:2408.06292 (2024).
\item
 Novikov A, Vũ N, Eisenberger M, et al. AlphaEvolve: A coding agent for scientific and algorithmic discovery. arXiv:2506.13131 (2025).
\item
 Lange RT, Imajuku Y, Cetin E. ShinkaEvolve: Towards Open-Ended and Sample-Efficient Program Evolution. arXiv:2509.19349 (2025).
\item
 Yang X, Yang X, Fang S, et al. R\&D-Agent: An LLM-Agent Framework Towards Autonomous Data Science. arXiv:2505.14738 (2025).
\item
 Zhang Y, Cheng X, Yuan M, et al. MoLRSI: A Recursive Self-Improving AI agent for Molecular Discovery. arXiv:2610.39714 (2026).
\item
 Pan Z, Feng Z, Cen Y, et al.~HarnessPro: Recursive Harness Self Improvement with Pareto Optimization for Performance-Cost Trade-offs. arXiv:2610.26770 (2026).
\item
 Barbaste P, Darrigol T, Vu G, Wiltberger T. Harness Engineering: Anatomy, Architecture, and Evolution of Coding Agents --- A Source-Code Study of Eleven Systems. arXiv:2609.00006 (2026).
\item
 Huang Y, Wang W, Bao H, et al.~MemoHarness: Agent Harnesses That Learn from Experience. arXiv:2607.14159 (2026).
\item
 Chan JS, Chowdhury N, Jaffe O, et al. MLE-bench: Evaluating Machine Learning Agents on Machine Learning Engineering. arXiv:2410.07095 (2024); \emph{International Conference on Learning Representations} (2025).
\item
 Kouno T, Silvas TV, Hilbert BJ, et al.~Crystal structure of APOBEC3A bound to single-stranded DNA reveals structural basis for cytidine deamination and specificity. \emph{Nature Communications} 8, 15024 (2017). \url{https://doi.org/10.1038/ncomms15024}
\item
 Jiang Y, Tang W, Sun H, et al. A Self-Evolving Agentic System for Automated Generation and Execution of Biological Protocols (ProtoPilot). arXiv:2606.31763 (2026).
\item
 Zhang C, Yang L, Huang L, et al. Self-evolving biofoundry for enzymatic engineering with traceable reasoning and agentic active learning (SEBio). Unpublished manuscript (2026).
\item
 Genoria AI Team, Hou Y, Yang Y, Tang W, Yang M. Operon-S1: a fast decision controller for long-horizon biomedical agents. arXiv:2610.41350 (2026).
\end{enumerate}

\par\endgroup

\phantomsection
\section*{Appendix}\label{ux9644ux5f55}
\addcontentsline{toc}{section}{Appendix}

\phantomsection
\subsection{A.1 Score records and plotting conventions}\label{a.1-ux5206ux6570ux8bb0ux5f55ux4e0eux4f5cux56feux53e3ux5f84}

\textbf{Score records and plotting.} All evaluation-by-evaluation scores were reconstructed by ordering on the remaining query quota in the host receipts and were checked item by item against the signed host receipts on the L40S. In Fig. 2 the horizontal axis counts the framework's own outer-loop rounds, including failed evaluations, the vertical axis is the cumulative best development-set PCC, and 0 is the official starting point. The native target for ShinkaEvolve and RD-Agent is 25 rounds, of which the first 20 are shown; the complete results are given in Table 3. The QwenCode curves in Figs 2 and 3 are taken throughout from the unmodified run reported in Table 3.

\phantomsection
\subsection{A.2 Supplementary notes on the evaluation resources}\label{a.2-ux8bc4ux6d4bux8d44ux6e90ux7684ux8865ux5145ux8bf4ux660e}

The 40 test environments of evolution-replay-v1 comprise 36 ProteinGym and 4 FLIP2 environments. The query protocol is 16 queries per round over 10 rounds, with the first round starting from WT plus 15 random points, across 5 seeds. Of the 2,401 saved replay records, 2,325 are development baselines and 76 are smoke tests.

\phantomsection
\subsection{A.3 Development-side reference baselines and the choice of statistical unit}\label{a.3-ux5f00ux53d1ux4fa7ux53c2ux8003ux57faux7ebfux4e0eux7edfux8ba1ux5355ux5143ux7684ux9009ux62e9}

On the development side of the budget-controlled replay suite (evolution-replay-v1), five reference strategies were each run over 93 development and validation environments $\times$ 5 seeds, for 2,325 replays in total. The size, splits and query protocol of this suite are given in Section 4.10 and Table 7. The normalised AUC of the best-discovery trajectory for each reference strategy is given in Table 1, in two columns: equal weight per related group, which is the primary basis, and equal weight per environment, shown as a sensitivity analysis:

\needspacefor{14.8\baselineskip}

\par\medskip\noindent{\small \textbf{Table 1 $|$ Normalised AUC of the best-discovery trajectory for the reference strategies on the development side of evolution-replay-v1.} 93 development and validation environments $\times$ 5 reference arms $\times$ 5 seeds, for 2,325 replays in total. Equal weight per related group is the primary basis and equal weight per environment is shown only as a sensitivity analysis.\par}\nopagebreak\vspace{-0.6\baselineskip}

\begingroup\small\setlength{\tabcolsep}{6pt}

\begin{longtable}[]{@{}
 >{\raggedright\arraybackslash}p{(\linewidth - 4\tabcolsep) * \real{0.3468}}
 >{\raggedright\arraybackslash}p{(\linewidth - 4\tabcolsep) * \real{0.3266}}
 >{\raggedright\arraybackslash}p{(\linewidth - 4\tabcolsep) * \real{0.3266}}@{}}
\toprule\noalign{}
\begin{minipage}[b]{\linewidth}\raggedright
Reference strategy
\end{minipage} & \begin{minipage}[b]{\linewidth}\raggedright
Equal weight per group (21 groups)
\end{minipage} & \begin{minipage}[b]{\linewidth}\raggedright
Equal weight per environment (93 env.)
\end{minipage} \\
\midrule\noalign{}
\endhead
\bottomrule\noalign{}
\endlastfoot
GP-BO & 0.6308 & 0.7388 \\
FolDE-style (reimplementation) & 0.6235 & 0.6684 \\
Static ESM2 & 0.6082 & 0.6782 \\
Linear ridge & 0.6063 & 0.7235 \\
Random & 0.5755 & 0.5669 \\
\end{longtable}

\endgroup

The two weightings do not give the same ranking: Linear ridge is second under equal weight per environment but falls to fourth under equal weight per group. This reversal follows directly from dependence between environments, because 64 of the 93 environments come from the same Tsuboyama group, and equal weight per environment therefore inflates the weight of that group to about two-thirds. The related group is accordingly used as the statistical unit throughout this work, with equal weight per environment shown only as a sensitivity analysis, and neither seeds nor assays from the same group may be treated as independent protein validation.

\phantomsection
\subsection{A.4 Tables}\label{a.4-ux8868}

\needspacefor{23.8\baselineskip}

\par\medskip\noindent{\small \textbf{Table 2 $|$ ThermoMPNN development-set results for the ten reference frameworks.} The official ThermoMPNN development-set PCC is 0.804358. Each framework was run for its own native number of outer-loop rounds, and one such round may be scored several times, so the number of scoring records can exceed the number of rounds.\par}\nopagebreak\vspace{-0.6\baselineskip}

\begingroup\scriptsize\setlength{\tabcolsep}{3pt}

\begin{longtable}[]{@{}
 >{\raggedright\arraybackslash}p{(\linewidth - 12\tabcolsep) * \real{0.1450}}
 >{\raggedright\arraybackslash}p{(\linewidth - 12\tabcolsep) * \real{0.1150}}
 >{\raggedright\arraybackslash}p{(\linewidth - 12\tabcolsep) * \real{0.0950}}
 >{\raggedright\arraybackslash}p{(\linewidth - 12\tabcolsep) * \real{0.1050}}
 >{\raggedright\arraybackslash}p{(\linewidth - 12\tabcolsep) * \real{0.1150}}
 >{\raggedright\arraybackslash}p{(\linewidth - 12\tabcolsep) * \real{0.1000}}
 >{\raggedright\arraybackslash}p{(\linewidth - 12\tabcolsep) * \real{0.3250}}@{}}
\toprule\noalign{}
\begin{minipage}[b]{\linewidth}\raggedright
Framework
\end{minipage} & \begin{minipage}[b]{\linewidth}\raggedright
Native outer-loop round
\end{minipage} & \begin{minipage}[b]{\linewidth}\raggedright
Done/target
\end{minipage} & \begin{minipage}[b]{\linewidth}\raggedright
Scored (valid)
\end{minipage} & \begin{minipage}[b]{\linewidth}\raggedright
Best DEV PCC
\end{minipage} & \begin{minipage}[b]{\linewidth}\raggedright
ΔPCC
\end{minipage} & \begin{minipage}[b]{\linewidth}\raggedright
Final state
\end{minipage} \\
\midrule\noalign{}
\endhead
\bottomrule\noalign{}
\endlastfoot
OpenRSI & native generation & 20/20 & 41 (40) & 0.834438 & +0.030080 & normal native return \\
AReaL-RSI & proposal experiment & 20/20 & 20 (20) & 0.831964 & +0.027606 & 20 completed; controller out of memory at wrap-up \\
AgenticEvolve & formal submission (2 islands) & 20/20 & 20 (19) & 0.823430 & +0.019072 & normal native return \\
ShinkaEvolve & native completions & 25/25 & 25 (24) & 0.813336 & +0.008979 & quota completed \\
MLEvolve & step attempt & 20/20 & 20 (15) & 0.812658 & +0.008301 & normal native return \\
RD-Agent & R\&D loop & 25/25 & 69 (69) & 0.811239 & +0.006881 & native return (across a recovery) \\
OpenEvolve & native iteration & 20/20 & 20 (16) & 0.807043 & +0.002685 & normal native return \\
TusoAI & terminated native job & 20/20 & 20 (15) & 0.806553 & +0.002196 & stopped by external budget \\
AI-Scientist-v3 & idea cycle & 4/20 & 19 (17) & 0.806209 & +0.001851 & stopped after a native version error \\
ERA & non-root expansion & 19/20 & 19 (14) & 0.804358 & +0.000000 & exited after cleanup failed at expansion 19 \\
\end{longtable}

\endgroup

Note: \(\Delta\)PCC is computed from full-precision values, so for some frameworks it differs by one unit in the last place from the difference of the rounded figures shown. The shape of the trajectory differs between frameworks. The main gain of OpenRSI comes from a single jump at outer-loop round 6 (0.8149 to 0.8319), and its final value was obtained from a 10-member ensemble of the official model plus 9 fully fine-tuned members. AReaL-RSI was advanced independently by 4 researchers on the same task, without selection or merging part-way through, and its best value comes from one of those lines. AgenticEvolve accumulated through many small increments. Most low-gain frameworks reached a plateau after the first few rounds.

\needspacefor{14.8\baselineskip}

\par\medskip\noindent{\small \textbf{Table 3 $|$ Run results for unmodified QwenCode and for EnzymeRSI.} The research code of both ThermoMPNN runs used ESM2 protein language model representations; the EnzymeRSI TadA run did not use ESM2.\par}\nopagebreak\vspace{-0.6\baselineskip}

\begingroup\scriptsize\setlength{\tabcolsep}{3pt}

\begin{longtable}[]{@{}
 >{\raggedright\arraybackslash}p{(\linewidth - 14\tabcolsep) * \real{0.1450}}
 >{\raggedright\arraybackslash}p{(\linewidth - 14\tabcolsep) * \real{0.1900}}
 >{\raggedright\arraybackslash}p{(\linewidth - 14\tabcolsep) * \real{0.1150}}
 >{\raggedright\arraybackslash}p{(\linewidth - 14\tabcolsep) * \real{0.0950}}
 >{\raggedright\arraybackslash}p{(\linewidth - 14\tabcolsep) * \real{0.0900}}
 >{\raggedright\arraybackslash}p{(\linewidth - 14\tabcolsep) * \real{0.0950}}
 >{\raggedright\arraybackslash}p{(\linewidth - 14\tabcolsep) * \real{0.1050}}
 >{\raggedright\arraybackslash}p{(\linewidth - 14\tabcolsep) * \real{0.1650}}@{}}
\toprule\noalign{}
\begin{minipage}[b]{\linewidth}\raggedright
Run
\end{minipage} & \begin{minipage}[b]{\linewidth}\raggedright
Task
\end{minipage} & \begin{minipage}[b]{\linewidth}\raggedright
Primary metric
\end{minipage} & \begin{minipage}[b]{\linewidth}\raggedright
Done/target
\end{minipage} & \begin{minipage}[b]{\linewidth}\raggedright
Valid evals
\end{minipage} & \begin{minipage}[b]{\linewidth}\raggedright
Best DEV
\end{minipage} & \begin{minipage}[b]{\linewidth}\raggedright
Wall clock (h)
\end{minipage} & \begin{minipage}[b]{\linewidth}\raggedright
Eval.\ index of first best
\end{minipage} \\
\midrule\noalign{}
\endhead
\bottomrule\noalign{}
\endlastfoot
Unmodified & ThermoMPNN & PCC & 20/20 & 20 & 0.839677 & 1.60 & 20 \\
EnzymeRSI & ThermoMPNN & PCC & 20/20 & 19 & 0.849462 & 2.54 & 20 \\
Unmodified & TadA amino-acid view & Spearman & 20/20 & 20 & 0.300716 & 2.24 & 18 \\
EnzymeRSI & TadA amino-acid view & Spearman & 20/20 & 20 & 0.306826 & 1.13 & 20 \\
\end{longtable}

\endgroup

\needspacefor{14.8\baselineskip}

\par\medskip\noindent{\small \textbf{Table 4 $|$ The five capabilities of the EnzymeRSI host-side research support layer.}\par}\nopagebreak\vspace{-0.6\baselineskip}

\begingroup\footnotesize\setlength{\tabcolsep}{4pt}

\begin{longtable}[]{@{}
 >{\raggedright\arraybackslash}p{(\linewidth - 6\tabcolsep) * \real{0.1109}}
 >{\raggedright\arraybackslash}p{(\linewidth - 6\tabcolsep) * \real{0.1417}}
 >{\raggedright\arraybackslash}p{(\linewidth - 6\tabcolsep) * \real{0.6519}}
 >{\raggedright\arraybackslash}p{(\linewidth - 6\tabcolsep) * \real{0.0955}}@{}}
\toprule\noalign{}
\begin{minipage}[b]{\linewidth}\raggedright
Class
\end{minipage} & \begin{minipage}[b]{\linewidth}\raggedright
Capability
\end{minipage} & \begin{minipage}[b]{\linewidth}\raggedright
Function
\end{minipage} & \begin{minipage}[b]{\linewidth}\raggedright
Quota
\end{minipage} \\
\midrule\noalign{}
\endhead
\bottomrule\noalign{}
\endlastfoot
Execution enhancement & Long background job & Training is submitted as an independent process group and is not bound by the 600 s limit on foreground commands; the agent polls the status and the tail of the log & Not consumed \\
Execution enhancement & Structured execution failure & Returns the exception type, the stage at which it failed and a one-line diagnosis, in place of an opaque transport error & Not consumed \\
Cross-framework integration & Research state query & Returns, from the signed host record, the current best, the full submission history including prediction hashes, the reasons for failure and the number of evaluations remaining & Not consumed \\
Cross-framework integration & Repeated-prediction pre-check & Returns a submission whose prediction vector is identical to an earlier one, and identifies the duplicated submission & Returns are not counted \\
Cross-framework integration & Delivery-manifest pre-check & Checks the submission list against the final packaging rules and returns it immediately if it does not comply & Returns are not counted \\
\end{longtable}

\endgroup

\needspacefor{11.2\baselineskip}

\par\medskip\noindent{\small \textbf{Table 5 $|$ The steps for which EnzymeRSI was responsible in the three-enzyme tests, and the nature of the evidence.}\par}\nopagebreak\vspace{-0.6\baselineskip}

\begingroup\footnotesize\setlength{\tabcolsep}{4pt}

\begin{longtable}[]{@{}
 >{\raggedright\arraybackslash}p{(\linewidth - 8\tabcolsep) * \real{0.0780}}
 >{\raggedright\arraybackslash}p{(\linewidth - 8\tabcolsep) * \real{0.2250}}
 >{\raggedright\arraybackslash}p{(\linewidth - 8\tabcolsep) * \real{0.2450}}
 >{\raggedright\arraybackslash}p{(\linewidth - 8\tabcolsep) * \real{0.1150}}
 >{\raggedright\arraybackslash}p{(\linewidth - 8\tabcolsep) * \real{0.3370}}@{}}
\toprule\noalign{}
\begin{minipage}[b]{\linewidth}\raggedright
Enzyme system
\end{minipage} & \begin{minipage}[b]{\linewidth}\raggedright
Measurement source
\end{minipage} & \begin{minipage}[b]{\linewidth}\raggedright
Steps EnzymeRSI performed
\end{minipage} & \begin{minipage}[b]{\linewidth}\raggedright
Evidence type
\end{minipage} & \begin{minipage}[b]{\linewidth}\raggedright
Main result
\end{minipage} \\
\midrule\noalign{}
\endhead
\bottomrule\noalign{}
\endlastfoot
RLuc & Luminescence measured through the cloud-laboratory workflow this round (a single round; initial screen plus the measurement of three independent preparations) & Nominated 40 single substitutions before measurement and froze the design record and evidence grading & \textbf{Prospective} & 10 of the 32 candidates not pre-specified reached $\geq$ 2$\times$ WT; all three inference-only candidates exceeded WT (2.30, 1.98 and 2.43$\times$), of which 2 reached at least 2$\times$; design rank was weakly correlated with measured rank (\(\text{ρ}\) = 0.175) \\
A3A & Dual-substrate conversion measured through the cloud-laboratory workflow for the variants nominated by EnzymeRSI (78 variants, 75 with complete pairs) & Generated the dual-context filtering and ranking rule and selected variants; analysed the measured results and performed structural analysis against 5KEG & Derived from measured variants & The selected variant changed between Y130Q, K47T and Y132C as the C limit was varied; Y130Q reversed the preference in both contexts; Tyr130 lies 3.66 \AA{} from the target C5 \\
KOD-E & Cy3-dATP incorporation efficiency measured through the cloud-laboratory workflow for 189 single-substitution variants & Paired parents with descendants, compared paths under two objectives and analysed epistasis; structural analysis was completed without access to activity values & Derived from measured variants & 125 single-step comparisons; a different objective gives a different next step; a reversal of rank appears at 676/680, where Ser minus His moves from $-1.9$ to $+5.5$ \\
\end{longtable}

\endgroup

\needspacefor{13.0\baselineskip}

\par\medskip\noindent{\small \textbf{Table 6 $|$ Tiered scoring and the corresponding decisions.}\par}\nopagebreak\vspace{-0.6\baselineskip}

\begingroup\footnotesize\setlength{\tabcolsep}{4pt}

\begin{longtable}[]{@{}
 >{\raggedright\arraybackslash}p{(\linewidth - 4\tabcolsep) * \real{0.1261}}
 >{\raggedright\arraybackslash}p{(\linewidth - 4\tabcolsep) * \real{0.5255}}
 >{\raggedright\arraybackslash}p{(\linewidth - 4\tabcolsep) * \real{0.3484}}@{}}
\toprule\noalign{}
\begin{minipage}[b]{\linewidth}\raggedright
Evaluation tier
\end{minipage} & \begin{minipage}[b]{\linewidth}\raggedright
Primary output
\end{minipage} & \begin{minipage}[b]{\linewidth}\raggedright
Corresponding decision
\end{minipage} \\
\midrule\noalign{}
\endhead
\bottomrule\noalign{}
\endlastfoot
Protein performance & Target activity, stability, selectivity and expression; difference relative to the reference and between replicates & Candidate priority, constraint satisfaction, next round of experiments \\
Algorithmic capability & Spearman correlation and PCC, top-\emph{k} hits, calibration, query curve, constraint violations & Method adoption, category of applicability, further development \\
System execution & Rate of valid completion, fault recovery, cost, latency, version and evidence integrity & Harness adoption, improvements to scheduling and workflow \\
Cross-task effect & Group-level paired gain, differences between independent runs, transfer and negative transfer & Global adoption, scoped retention, rollback \\
\end{longtable}

\endgroup

\needspacefor{18.4\baselineskip}

\par\medskip\noindent{\small \textbf{Table 7 $|$ Tiered evaluation resources and the conclusions each tier can support.} Every split is frozen by sha256 in its own \texttt{manifest.json}, and the statistical unit is the related group.\par}\nopagebreak\vspace{-0.6\baselineskip}

\begingroup\footnotesize\setstretch{1.0}\setlength{\tabcolsep}{4pt}

\begin{longtable}[]{@{}
 >{\raggedright\arraybackslash}p{(\linewidth - 8\tabcolsep) * \real{0.0856}}
 >{\raggedright\arraybackslash}p{(\linewidth - 8\tabcolsep) * \real{0.1498}}
 >{\raggedright\arraybackslash}p{(\linewidth - 8\tabcolsep) * \real{0.2623}}
 >{\raggedright\arraybackslash}p{(\linewidth - 8\tabcolsep) * \real{0.2191}}
 >{\raggedright\arraybackslash}p{(\linewidth - 8\tabcolsep) * \real{0.2831}}@{}}
\toprule\noalign{}
\begin{minipage}[b]{\linewidth}\raggedright
Tier
\end{minipage} & \begin{minipage}[b]{\linewidth}\raggedright
Suite
\end{minipage} & \begin{minipage}[b]{\linewidth}\raggedright
Size and splits
\end{minipage} & \begin{minipage}[b]{\linewidth}\raggedright
Conclusions carried
\end{minipage} & \begin{minipage}[b]{\linewidth}\raggedright
Boundary not to be crossed
\end{minipage} \\
\midrule\noalign{}
\endhead
\bottomrule\noalign{}
\endlastfoot
A Static task & \texttt{evolution-\allowbreak{}v1} & 217 assays mapped to 101 groups; the outer loop takes one assay from each of five classes of measurement and self-evolves assay by assay, for 196 rounds in total & The five endpoints ranked alongside the 97 published models of the official ProteinGym record; the task base for H0/H1 paired comparison & Self-improvement assay by assay does not demonstrate generalisation across assays; formal attribution still requires the same algorithm assets, base model, environment and budget, together with traceable version changes and adoption records \\
A Budget replay & \texttt{evolution-\allowbreak{}replay-\allowbreak{}v1} & 133 environments; 88 development, 5 validation, 40 test (36 ProteinGym and 4 FLIP2); 16 queries $\times$ 10 rounds $\times$ 5 seeds & Quality of sequential decisions and query efficiency for the five reference strategies (Appendix A.3); no system arm is reported on this tier in the present work & Selecting candidates over several rounds does not by itself constitute algorithm RSI, which requires evidence of a persistent change in the method artefact together with an equal-budget HPO control \\
B Real-time extrapolation & \texttt{TadA-Bench} & 31 rounds of PANCE; training rounds 1--27, validation round 28, test rounds 29--31 (sealed) & The ability to extrapolate the next round from past experiments & The DNA, RNA and amino-acid views come from the same experimental source and may not be counted as three independent campaigns; the data must never be re-split at random \\
B-mini Closed-loop campaign & \texttt{external-\allowbreak{}campaign-\allowbreak{}v1} & AtHMT, YmPhytase and MjTyrRS; 1,321 assayed variants; 5, 4 and 4 rounds; the final round sealed & Ranking the next-round candidates that the authors actually selected & This may not be read as discovering candidates in an unconstrained design space; it serves as supplementary evidence only \\
B Supplementary & \texttt{external-\allowbreak{}pance-\allowbreak{}v1} & POGO-PANCE single lineage R2\_P1-2; 569 genotypes, 22 hotspots, 3 cycles; the final cycle sealed & Cycle-level prediction of the selection outcome of the next round & A selection outcome measured as pool frequency is not equivalent to the activity of the purified enzyme; a single lineage serves as supplementary evidence only \\
C External transfer & \texttt{external-\allowbreak{}mavedb-\allowbreak{}v1} & 12 assays in 12 groups; 105,783 variants in the cumulative metadata & Analysis of transfer across proteins & Being a different protein from those in ProteinGym does not by itself imply the absence of homology, the absence of exposure during pre-training, or generalisation to an entirely new task \\
C Small cohort & \texttt{external-\allowbreak{}evolvepro-\allowbreak{}v1} & 6 protein tasks; 481 rows, 422 assayed, 419 entering the denominator; evaluation only & Zero-shot ranking (Spearman correlation, top 10\%, nDCG@10) & A flat cohort with no complete round structure; duplicate records do not increase the number of independent proteins \\
\end{longtable}

\endgroup

\end{document}